\documentclass[authoryear,3p,review,12pt]{elsarticle}

\usepackage{amssymb}
\usepackage{amsmath,amsthm}
\usepackage{cancel}
\usepackage{graphicx,helvet}
\usepackage{subfigure,hyperref,multirow}
\usepackage[ruled,vlined]{algorithm2e}
\usepackage{booktabs}
\newtheorem{property}{Property}

\begin{document}

\begin{frontmatter}



\title{An Enhanced Branch-and-bound Algorithm for the Talent Scheduling Problem}

\author[cityu]{Zizhen Zhang}
\ead{zhangzizhen@gmail.com}

\author[hust]{Hu Qin\corref{corl}}
\ead{tigerqin@hust.edu.cn, tigerqin1980@gmail.com}

\author[sysu]{Xiaocong Liang}
\ead{yurilxc@gmail.com}

\author[cityu]{Andrew Lim}
\ead{lim.andrew@cityu.edu.hk}

\address[cityu]{
Department of Management Sciences,
City University of Hong Kong,\\
Tat Chee Ave., Kowloon Tong, Kowloon, Hong Kong
}

\address[hust]{
School of Management, Huazhong University of Science and Technology,\\
No. 1037, Luoyu Road, Wuhan, China
}

\address[sysu]{
Department of Computer Science, School of Information Science and Technology,\\
Sun Yat-Sen University, Guangzhou, China \\
}

\cortext[corl]{Corresponding author at: School of Management, Huazhong University of
Science and Technology, No. 1037, Luoyu Road, Wuhan, China. Tel.: +86
13349921096.
}

\begin{abstract}
The talent scheduling problem is a simplified version of the real-world film shooting problem, which aims to determine a shooting sequence so as to minimize the total cost of the actors involved. In this article, we first formulate the problem as an integer linear programming model. Next, we devise a branch-and-bound algorithm to solve the problem. The branch-and-bound algorithm is enhanced by several accelerating techniques, including preprocessing, dominance rules and caching search states. Extensive experiments over two sets of benchmark instances suggest that our algorithm is superior to the current best exact algorithm. Finally, the impacts of different parameter settings are disclosed by some additional experiments.
\end{abstract}


\end{frontmatter}

\section{Introduction}
\label{sec:i}
The scenes of a film are not generally shot in the same sequence as they appear in the final version. Finding an optimal sequence in which the scenes are shot motivates the investigation of the talent scheduling problem, which is formally described as follows. Let $S=\{s_1, s_2, \ldots, s_n\}$ be a set of $n$ scenes and $A=\{a_1, a_2, \ldots, a_m\}$ be a set of $m$ actors. All scenes are assumed to be shot on a given location. Each scene $s_j \in S$ requires a subset $a(s_j) \subseteq A$ of actors and has a duration $d(s_j)$ that commonly consists of one or several days. Each actor $a_i$ is required by a subset $s(a_i)\subseteq S$ of scenes. We denote by $\Pi$ the permutation set of the $n$ scenes and define $e_{i}(\pi)$ (respectively, $l_{i}(\pi)$) as the earliest day (respectively, the latest day) in which actor $i$ is required to be present on location in the permutation $\pi \in \Pi$. Each actor $a_i \in A$ has a daily wage $c(a_i)$ and is paid for each day from $e_{i}(\pi)$ to $l_{i}(\pi)$ regardless of whether they are required in the scenes. The objective of the talent scheduling problem is to find a shooting sequence (i.e., a permutation $\pi \in \Pi$) of all scenes that minimizes the total paid wages.

Table \ref{tab:1} presents an example of the talent scheduling problem, which is reproduced from \citet{de2011solving}. The information of $a(s_j)$ and $s(a_i)$ is determined by the {$m \times n$} matrix $M$ shown in Table \ref{tab:1}(a), where cell $M_{i, j}$ is filled with an ``$X$'' if actor $a_i$ participates in scene $s_j$ and with a ``$\cdot$'' otherwise. Obviously, we can obtain $a(s_j)$ and $s(a_i)$ by $a(s_j) = \{a_i| M_{i, j} = X\}$ and $s(a_i) = \{s_j|M_{i, j} = X\}$, respectively. The last row gives the duration of each scene and the rightmost column gives the daily cost of each actor. If the shooting sequence is $\pi = \{s_1, s_2, s_3, s_4, s_5, s_6, s_7, s_8, s_9, s_{10}, s_{11}, s_{12} \}$, we can get a matrix $M(\pi)$ shown in Table \ref{tab:1}(b), where in cell $M_{i, j}(\pi)$ a sign ``$X$'' indicates that actor $a_i$ participates in scene $s_j$ and a sign ``--'' indicates that actor $a_i$ is waiting at the filming location. The cost of each scene is presented in the second-to-last row and the total cost is 604. The cost incurred by the waiting status of the actors is called {\em holding cost}, which is shown in the last row of Table \ref{tab:1}(b). The optimal solution of this instance is $\pi^* = \{s_5, s_2, s_7, s_1, s_6, s_8, s_4, s_9, s_3, s_{11}, s_{10}, s_{12} \}$ whose total cost and holding cost are 434 and 53, respectively.
\begin{table*}[!h]
  \scriptsize
  \centering
  \caption{An example of the talent scheduling problem reproduced from \citet{de2011solving}.}
  \label{tab:1}
  \subtable[The matrix $M$ for an instance of the talent scheduling problem.] {
    \begin{tabular}{c|cccccccccccc|c}
    \hline
          & $s_1$    & $s_2$    & $s_3$    & $s_4$    & $s_5$    & $s_6$    & $s_7$    & $s_8$    & $s_9$    & $s_{10}$   & $s_{11}$   & $s_{12}$   & $c(a_i)$ \\
    \hline
    $a_1$    &$\mathrm{X}$   & $\cdot$     & X     & $\cdot$     & $\cdot$     & X     & $\cdot$     & X     & X     & X     & X     & X     & 20 \\
    $a_2$    & X     & X     & X     & X     & X     & $\cdot$    & X     & $\cdot$     & X     & $\cdot$     & X     & $\cdot$     & 5 \\
    $a_3$    & $\cdot$     & X     & $\cdot$     & $\cdot$     & $\cdot$     & $\cdot$     & X     & X     & $\cdot$     & $\cdot$    & $\cdot$     & $\cdot$     & 4 \\
    $a_4$    & X     & X     & $\cdot$     & $\cdot$    & X     & X     & $\cdot$     & $\cdot$     & $\cdot$     & $\cdot$     & $\cdot$     & $\cdot$     & 10 \\
    $a_5$    & $\cdot$     & $\cdot$    & $\cdot$     & X     & $\cdot$    & $\cdot$     & $\cdot$     & X     & X     & $\cdot$     & $\cdot$     & $\cdot$     & 4 \\
    $a_6$    & $\cdot$    & $\cdot$    & $\cdot$   & $\cdot$     & $\cdot$     & $\cdot$     & $\cdot$     & $\cdot$     & $\cdot$     & X     & $\cdot$     & $\cdot$     & 7 \\
    \hline
    $d(s_j)$  & 1     & 1     & 2     & 1     & 3     & 1     & 1     & 2     & 1     & 2     & 1     & 1     &  \\
    \hline
    \end{tabular}
   	\label{tab:11}
  }
  \subtable[The matrix $M(\pi)$ corresponding to a solution $\pi$ of the instance.] {
    \begin{tabular}{c|cccccccccccc|c}
    \hline
          & $s_1$    & $s_2$    & $s_3$    & $s_4$    & $s_5$    & $s_6$    & $s_7$    & $s_8$    & $s_9$    & $s_{10}$   & $s_{11}$   & $s_{12}$   & $c(a_i)$ \\
    \hline
    $a_1$    & X     & --     & X     & --     & --     & X     & --     & X     & X     & X     & X     & X     & 20 \\
    $a_2$    & X     & X     & X     & X     & X     & --     & X     & --     & X     & --     & X     & $\cdot$     & 5 \\
    $a_3$    & $\cdot$    & X     & --     & --     & --     & --     & X     & X     & $\cdot$     & $\cdot$     & $\cdot$     & $\cdot$    & 4 \\
    $a_4$    & X     & X     & --     & --     & X     & X     & $\cdot$    & $\cdot$     & $\cdot$    & $\cdot$    & $\cdot$    & $\cdot$     & 10 \\
    $a_5$    &$\cdot$     &$\cdot$    & $\cdot$    & X     & --     & --     & --     & X     & X     & $\cdot$     & $\cdot$     & $\cdot$     & 4 \\
    $a_6$    & $\cdot$     & $\cdot$     & $\cdot$    & $\cdot$     & $\cdot$     & $\cdot$    & $\cdot$    & $\cdot$     & $\cdot$   & X     & $\cdot$     & $\cdot$     & 7 \\
    \hline
    cost  & 35     & 39     & 78     & 43     & 129     & 43     & 33     & 66     & 29     & 64     & 25     & 20     & 604 \\
    holding cost & 0     & 20     & 28     & 34     & 84     & 13     & 24     & 10     & 0     & 10     & 0     & 0     & 223 \\
    \hline
    \end{tabular}
  	\label{tab:12}
   }
\end{table*}

The talent scheduling problem was originated from \citet{adelson1976dynamic} and \citet{cheng1993optimal}. \citet{adelson1976dynamic} introduced an orchestra rehearsal scheduling problem, which can be viewed as a restricted version of the talent scheduling problem with all actors having the same daily wage. They proposed a simple dynamic programming algorithm to solve their problem. \citet{cheng1993optimal} studied a film scheduling problem in which all scenes have identical duration. They first showed that the problem is NP-hard even if each actor is required by two scenes and the daily wage of each actor is one. Next, they devised a branch-and-bound algorithm and a simple greedy hill climbing heuristic to solve their problem. Later, \citet{smith2003constraint} applied constraint programming to solve both the problems introduced by \citet{adelson1976dynamic} and \citet{cheng1993optimal}. In her subsequent work, namely \citet{smith2005caching}, she accelerated her constraint programming approach by caching search states.

The talent scheduling problem we study in this article was first formally described by \citet{de2011solving}. This problem is a generalization of the problems introduced by \citet{adelson1976dynamic} and \citet{cheng1993optimal}, where scenes may have different durations and actors may have different wages. However, it is a simplified version of the movie shoot scheduling problem (MSSP) introduced by \citet{bomsdorf2008model}. In the MSSP, we need to deal with a couple of practical constraints, such as the precedence relations among scenes, the time windows of each scene, the resource availability, and the working time windows of actors and other film crew members.

 %To the best of our knowledge, the dynamic programming algorithm developed by \citet{de2011solving} is the current best exact algorithm for the talent scheduling problem.

In literature, there exist several meta-heuristics developed for the problem introduced by \citet{cheng1993optimal}. \citet{nordstrom1994genetic} provided several hybrid genetic algorithms for this problem and showed that their algorithms outperform the heuristic approach in \citet{cheng1993optimal} in terms of both solution quality and computation speed. \citet{fink1999applications} treated this problem as a special application of the general pattern sequencing problem, and implemented a simulated annealing algorithm and several tabu search heuristics to solve it.

The talent scheduling problem is a very challenging combinatorial optimization problem. The current best exact approach by \citet{de2011solving} can only optimally solve small- and medium-size instances. In this paper, we propose an enhanced branch-and-bound algorithm for the talent scheduling problem, which uses the following two main techniques:
\begin{itemize}
\item {\em Dominance rules.} When a partial solution represented by a node in the search tree can be dominated by another partial solution, this node need not be further explored and can be safely discarded.
\item {\em Caching search states.} The talent scheduling problem can be solved by dynamic programming algorithm (see \citet{de2011solving}). It is beneficial to incorporate the dynamic programming states into the branch-and-bound framework by memoization technique. In the branch-and-bound tree, each node is related to a dynamic programming state. If the search process explores a certain node whose already confirmed cost is not smaller than the value of its corresponding cached state, this node can be pruned.
\end{itemize}

There are three main contributions in this paper. Firstly, we formulate the talent scheduling problem as a mixed integer linear programming model so that commercial mathematical programming solvers can be applied to the problem. Secondly, we propose an enhanced branch-and-bound algorithm whose novelties include a new lower bound, caching search states and two problem-specific dominance rules. Thirdly, we achieved the optimal solutions for more benchmark instances by our algorithm. The experimental results show that our branch-and-bound algorithm is superior to the current best exact approach by \citet{de2011solving}.

The remainder of this paper is organized as follows. In Section \ref{sec:mf}, we present the mixed integer linear programming model for the talent scheduling problem. Next, we describe our branch-and-bound algorithm in Section \ref{sec:app}, including the details on a double-ended search strategy, the computation of the lower bound, a preprocessing step, the state caching process and the dominance rules. The computational results are reported in Section \ref{sec:exp}, where we use our algorithm to solve over 200,000 benchmark instances. Finally, we conclude our study in Section \ref{sec:con} with some closing remarks.

\section{Mathematical Formulation}
\label{sec:mf}
The talent scheduling problem is essentially a permutation problem. It tries to find a permutation (i.e., a schedule) $\pi =(\pi(1), \ldots, \pi(n)) \in \Pi$, where $\pi(k)$ is the $k$-th scene in permutation $\pi$, such that the total cost $C(\pi)$ is minimized. The value of $C(\pi)$ is computed as:
\begin{small}
\begin{align}
C(\pi)=\sum_{i = 1}^{m} c(a_i)\times \big{(}l_i(\pi) - e_i(\pi) + 1 \big{)} \nonumber
\end{align}
\end{small}
We set the parameter $m_{i,j} =1$ if $M_{i, j} = X$ and $m_{i, j} = 0$ otherwise. The total holding cost can be easily derived as:
\begin{small}
\begin{align}
H(\pi)=\sum_{ i = 1}^{m} c(a_i) \times \Big{(}l_i(\pi) - e_i(\pi) + 1 - \sum_{j = 1}^{n} m_{i,j}d(s_j)\bigg{)} \nonumber
\end{align}
\end{small}
Apparently, for this problem minimizing the total cost is equivalent to minimizing the total holding cost.

The talent scheduling problem can be formulated into an integer linear programming formulation using the following decision variables:
\begin{description}
\item $x_{i,j}$: a binary variable that equals 1 if scene $s_j$ is scheduled immediately after scene $s_i$, and 0 otherwise.
\item $t_j$: the starting day for shooting scene $s_j$.
\item $e_i$: the earliest shooting day that requires actor $a_i$.
\item $l_i$: the latest shooting day that requires actor $a_i$.
\end{description}

The integer programming formulation is given by:
\begin{small}
\begin{align}
(\textrm{IP})~~& \min~~\sum_{i = 1}^{m} c(a_i)(l_i - e_i + 1) \label{eqn:obj} \\
\mbox{s.t.}~ & \sum_{ j = 0, i \neq j}^{n} x_{i, j} = 1, ~ \forall ~ 0 \leq i \leq n \label{eqn:c1} \\
	& \sum_{ i = 0, i \neq j}^{n} x_{i, j} = 1, ~ \forall~ 0 \leq j \leq n \label{eqn:c2} \\
	& \sum_{j=0, i \neq j}^{n} t_j x_{i, j} = t_i + d(s_i), ~ \forall~ 1 \leq i \leq n \label{eqn:c3} \\
	& e_i \leq t_j, ~ \forall~ 1 \leq i \leq m, ~ 1 \leq j \leq n, ~ m_{i,j}=1 \label{eqn:c4} \\
	& t_j + d(s_j) - 1 \leq l_i, ~ \forall~ 1 \leq i \leq m, ~ 1 \leq j \leq n, ~ m_{i,j} = 1 \label{eqn:c5} \\
	& x_{i, j} \in \{0, 1\}, ~ \forall~ 1 \leq i, ~ j \leq n \label{eqn:c6} \\
	& e_i, l_i, t_j \geq 0~ \textrm{and integer}, ~ \forall~ 1 \leq i \leq m,~ 1 \leq j \leq n \label{eqn:c7}
\end{align}
\end{small}
The objective (\ref{eqn:obj}) is to minimize the total cost, where $l_i - e_i + 1$ is the number of days in which actor $a_i$ is present on location. Constraints (\ref{eqn:c1}) and (\ref{eqn:c2}) guarantee that every scene has exactly one immediate successor and one immediate predecessor, respectively. Note that we create a dummy scene $s_0$ which enables us to identify the first and the last scene to be shot. Constraints (\ref{eqn:c3}) state that the starting day of scene $s_j$ is determined by the starting day of its predecessor scene $s_i$. From this set of constraints, we can conclude that the starting day of the dummy scene $s_0$ equals $\sum_{j = 1}^{n} d(s_j)$. Moreover, these constraints prevent sub-tours from occurring. Constraints (\ref{eqn:c4}) and  (\ref{eqn:c5}) ensure that the earliest and the latest shooting days that require actor $a_i$ are determined by the starting days of scenes in which he/she is involved.

Observe that Constraints (\ref{eqn:c3}) are nonlinear. To linearize them, we introduce a set of additional variables $z_{i, j} (1 \leq i \leq n, 0 \leq j \leq n, i \neq j)$, and set $z_{i,j} = t_j x_{i,j}$. We know that $z_{i, j} = t_j$ if $x_{i, j} = 1$ and $z_{i,j} = 0$ otherwise. Thus, $z_{i,j}$ can be restricted by the following four linear constraints:
\begin{small}
\begin{align}
z_{i, j} \geq 0	\label{eqn:c8} \\
z_{i, j} \leq t_j	\label{eqn:c9}	\\
z_{i, j} \geq t_j + L (x_{i,j}-1)	\label{eqn:c10}	\\
z_{i, j} \leq L x_{i,j}	\label{eqn:c11}
\end{align}
\end{small}
where $L$ is a sufficiently large positive number, e.g., $L = \sum_{j=1}^{n} d(s_j)$. Accordingly, Constraints (\ref{eqn:c3}) can be rewritten as:
\begin{small}
\begin{align}
	& \sum_{j = 0, i \neq j}^{n}z_{i, j} = t_i + d(s_i), ~ \forall~ 1 \leq i \leq n \label{eqn:cc3}
\end{align}
\end{small}
The objective (\ref{eqn:obj}) and Constraints (\ref{eqn:c1}) -- (\ref{eqn:c2}), (\ref{eqn:c4}) -- (\ref{eqn:cc3}) constitute an integer linear programming model (ILP) for the talent scheduling problem. This ILP is quite difficult to be optimally solved by commercial integer programming solvers, e.g., ILOG CPLEX. Preliminary experiments revealed that only very small-scale instances, e.g., $n=10$ and $m=5$, can be optimally solved by CPLEX 12.1 with default settings. This is mainly because the linear relaxation of the ILP model cannot provide a high-quality lower bound for the problem.

\section{An Enhanced Branch-and-bound Approach}
\label{sec:app}
Branch-and-bound is a general technique for optimally solving various combinatorial optimization problems. The basic idea of the branch-and-bound algorithm is to systematically and implicitly enumerate all candidate solutions, where large subsets of fruitless candidates are discarded by using upper and lower bounds, and dominance rules. In this section, we describe the main components of our proposed branch-and-bound algorithm, including a double-ended search strategy, a novel lower bound, the preprocessing stage, the state caching strategy and two dominance rules. For the rest of this discussion, we choose minimizing the total holding cost as the objective of the talent scheduling problem.

\subsection{Double-ended Search}
\label{sec:ss}
The solutions of the talent scheduling problem can be easily presented in a branch-and-bound search tree. Suppose we aim to find an optimal permutation $\pi^* $ $=(\pi^*(1),$ $\pi^*(2),$ $\ldots,$ $\pi^*(n))$. A typical branch-and-bound process first determines the first $k$ scenes to be shot, denoted by a partial permutation $(\hat{\pi}(1), \ldots, \hat{\pi}(k))$, at level $k$ of the search tree. Then, it generates $n - k$ branches, each trying to explore a node by assigning a scene to $\pi(k + 1)$. At some tree node at level $k+1$, there is a known partial permutation $(\hat{\pi}(1), \hat{\pi}(2), \ldots, \hat{\pi}(k+1))$ and a set of $n -  k - 1$ unscheduled scenes. If the lower bound $LB$ to the value of the solutions that contain the partial permutation $(\hat{\pi}(1), \hat{\pi}(2), \ldots, \hat{\pi}(k+1))$ is not less than the current best solution value (i.e., an upper bound $UB$), then the branch to the node associated with $\hat{\pi}(k+1)$ can be safely discarded. Once the search process reaches a node at level $n$ of the tree, a feasible solution is obtained and the current best solution may be updated accordingly.

The above search methodology can be called the {\em single-ended search strategy}. As did by \citet{cheng1993optimal} and \citet{de2011solving}, we can employ a {\em double-ended search strategy} that alternatively fixes the first and the last undetermined positions in the permutation. That is to say, the double-ended search determines a scene permutation following the order $\pi(1), \pi(n), \pi(2), \pi(n-1)$ and so on. When using the double-ended search strategy, a node in some level of the search tree corresponds to a partially determined permutation with the form $(\hat{\pi}(1), \ldots, \hat{\pi}(k-1), \pi(k), \ldots, \pi(l), \hat{\pi}(l+1), \ldots, \hat{\pi}(n))$, where $1\leq k\leq l\leq n$ and the value of $\pi(h)$ $(k \leq h \leq l)$ is undetermined. We denote by $B$ the set of scenes scheduled at the beginning of the permutation, namely $B = \{\hat{\pi}(1), \hat{\pi}(2), \ldots, \hat{\pi}(k-1)\}$, and by $E$ the set of scenes scheduled at the end, namely $E = \{\hat{\pi}(l+1), \hat{\pi}(l+2), \ldots, \hat{\pi}(n)\}$. The remaining scenes are put in a set $Q$, namely $Q = S - B - E$. Moreover, for convenience, we denote by $\vec{B}$ and $\vec{E}$ the partially determined scene sequences at the beginning and at the end of a permutation, i.e., $\vec{B} = (\hat{\pi}(1), \ldots, \hat{\pi}(k-1))$ and $\vec{E} = (\hat{\pi}(l+1), \ldots, \hat{\pi}(n))$.

The double-ended search strategy is beneficial to solving the talent scheduling problem. As pointed out by \citet{de2011solving}, it can help obtain more accurate lower bounds by increasing the number of fixed actors. The actor required by the scenes in both $B$ and $E$ is labeled {\em fixed} since the total number of his/her on-location days is fixed and his/her cost in the final schedule already becomes known. We do not need to consider any fixed actor in the later stages of the search process, which certainly reduces the size of the problem. Let $a(Q) = \cup_{s\in Q}a(s)$ be the set of actors required by at least one scene in $Q \subseteq S$. The set of fixed actors can be represented by $F = a(B) \cap a(E)$.

A generic double-ended branch-and-bound framework is given in Algorithm \ref{alg:1}. The operator ``$\circ$'' in lines \ref{algo1:line:2} and \ref{algo1:line:13} indicates concatenating two partially determined scene sequences. The function $\textbf{search}(\vec{B}, Q, \vec{E})$ returns the optimal solution to the talent scheduling problem with known $\vec{B}$ and $\vec{E}$, denoted by $P(\vec{B}, Q, \vec{E})$. The optimal solution of the talent scheduling problem can be achieved by invoking $\textbf{search}(\vec{B}, Q, \vec{E}$) with $B = E = \emptyset$ and $Q = S$.  The function {\bf evaluate}({\em solution}) returns the value of {\em solution}. The function $\textbf{lower\_bound}(\vec{B} \circ s, Q - \{s\}, \vec{E})$ provides a valid lower bound to problem $P(\vec{B}\circ s, Q - \{s\}, \vec{E})$, where the set $B$ of scenes is scheduled before scene $s$ and the set $S - B - \{s\}$ of scenes is scheduled after scene $s$. The branch-and-bound search tries to schedule each remaining scene $s$ immediately after $\vec{B}$, and then swaps the roles of $\vec{B}$ and $\vec{E}$ to continue building the search tree (see line \ref{algo1:line:13}).

\begin{algorithm}
\SetAlgoNoLine
\LinesNumbered
\SetCommentSty{textit}
\small
\SetKwFunction{search}{search}
\SetKwInput{Func}{Function} \BlankLine
\Func{$\textbf{search}(\vec{B}, Q, \vec{E}$)}
\If {$Q = \emptyset$} {
	$current\_solution=\vec{B}\circ \vec{E}$\; \label{algo1:line:2}
	$z=\textbf{evaluate}(current\_solution)$\;
	\If {$z < UB$} {
		$UB: = z$\;
		$best\_solution:=current\_solution$\;
	}
	\Return\;
}
\ForEach {$s \in Q$} {
	$LB:=\textbf{lower\_bound}(\vec{B} \circ s, Q - \{s\}, \vec{E})$\;
	\lIf {$LB\geq UB$} continue\;
	$\textbf{search}(\vec{E}, Q- \{s\}, \vec{B} \circ s)$\; \label{algo1:line:13}
}
\caption{A generic double-ended branch-and-bound search framework.}
\label{alg:1}
\end{algorithm}

\subsection{Lower Bound to $P(\vec{B}, Q, \vec{E})$}
\label{sec:dec}
The problem $P(\vec{B}, Q, \vec{E})$ corresponds to a node in the search tree. Its lower bound $\textbf{lower\_bound}(\vec{B}, Q, \vec{E})$ can be expressed as:
\begin{small}
\begin{align}
\textbf{lower\_bound}(\vec{B}, Q, \vec{E}) = cost(\vec{B}, \vec{E}) + \textbf{lower}(B, Q, E), \nonumber
%\label{eqn:x1}
\end{align}
\end{small}
where $cost(\vec{B}, \vec{E})$, called {\em past cost}, is the cost incurred by the path from the root node to the current node, and $\textbf{lower}(B, Q, E)$ provides a lower bound to {\em future cost}, i.e., the holding cost to be incurred by scheduling the scenes in $Q$. We discuss the past cost $cost(\vec{B}, \vec{E})$ in this subsection and leave the description of $\textbf{lower}(B, Q, E)$ in Subsection \ref{sec:lb}.

When $\vec{B}$ and $\vec{E}$ have been fixed, a portion of holding cost, namely $cost(\vec{B}, \vec{E})$, is determined regardless of the schedule of the scenes in $Q$.
The past cost $cost(\vec{B}, \vec{E})$ is incurred by the holding days that can be confirmed by the following three ways:
\begin{enumerate}
\item For the actor $a_i \in a(B) \cap a(E)$, the number of his/her holding days in any complete schedule can be fixed \citep{cheng1993optimal}.
\item For the actor $a_i \in a(B) \cap a(Q) - a(E)$, the number of his/her holding days in the time period for completing scenes in $B$ can be fixed.
\item For the actor $a_i \in a(E) \cap a(Q) - a(B)$, the number of his/her holding days in the time period for completing scenes in $E$ can be fixed.
\end{enumerate}

Furthermore, we use $cost(s, B, E)$ to represent the newly confirmed holding cost incurred by placing scene $s \in Q$ at the first unscheduled position, namely the position after any scene in $B$ and before any scene in $S - B - \{s\}$. Note that $cost(s, B, E)$ is irrelevant to the orders of scenes in $B$ and $E$. Obviously, we have $cost(\vec{B} \circ \{s\}, \vec{E}) =  cost(\vec{B}, \vec{E}) + cost(s, B, E)$, which implies that the past cost of a tree node is the sum of the past cost of its father node and the newly confirmed holding cost incurred by branching. As a result, the lower bound function can be rewritten as:
\begin{small}
\begin{align}
\textbf{lower\_bound}(\vec{B} \circ s, Q - \{s\}, \vec{E}) = cost(\vec{B}, \vec{E}) + cost(s, B, E)
 +\textbf{lower}(B\cup\{s\},Q-\{s\},E). \nonumber
%\label{eqn:lb}
\end{align}
\end{small}

The value of $cost(s, B, E)$ is incurred by the following two type of actors:

{\em Type 1.} If actor $a_i$ is included in neither $a(B) \cap a(E)$ nor $a(s)$ but is still present on location during the days of shooting scene $s$ (i.e., $a_i \notin a(B) \cap a(E)$, $a_i \notin a(s)$ and $a_i \in a(B) \cap a(Q-\{s\})$), he/she must be held during the shooting days of scene $s$.

{\em Type 2.} If actor $a_i$ is not included in $a(B) \cap a(E)$ but is included in $a(E)$, and  scene $s$ is his/her first involved scene (i.e., $a_i \notin a(B)$ and $a_i \in a(s)$ and $a_i \in a(E)$), the shooting days of those scenes in $Q - \{s\}$ that do not require actor $a_i$ can be confirmed as his/her holding days.

To demonstrate the computation of $cost(\vec{B}, \vec{E})$ and $cost(s, B, E)$, let us consider a partial schedule presented in Table \ref{tab:z}, where $\vec{B} = (s_1, s_2)$, $\vec{E} = (s_5,s_6)$ and $Q = S - B - E = \{s_3, s_4\}$. In the columns ``$cost(\vec{B}, \vec{E})$'', ``$cost(s_3, B, E)$'' and ``$cost(s_4, B, E)$'', we present the corresponding holding cost associated with each actor. For example, the value of $cost(\vec{B}, \vec{E})$ can be obtained by summing up the values in all cells of the column ``$cost(\vec{B}, \vec{E})$''. Since actor $a_1$ is a fixed actor, his/her holding cost must be $c(a_1)(d(s_2) + d(s_4))$ no matter how the scenes in $Q$ are scheduled. Actor $a_2$ is involved in $B$ and $Q$ but is not involved in $E$, so we can only say that the holding cost of this actor is at least $c(a_2)d(s_2)$. Similarly, actor $a_3$ has an already incurred holding cost $c(a_3)d(s_5)$. For actors $a_4$ and $a_5$, we cannot get any clue on their holding costs from this partial schedule and thus we say their already confirmed holding costs are both zero. Suppose scene $s_4$ is placed at the first unscheduled position. Since actors $a_2$ and $a_4$ must be present on location during the period of shooting scene $s_4$, the newly confirmed holding cost is $ cost(s_4, B, E) = (c(a_2) + c(a_4))d(s_4)$. If we suppose scene $s_3$ is placed at the first unscheduled position, the newly confirmed holding cost is only related to actor $a_3$, namely, $cost(s_3, B, E) = c(a_3)d(s_4)$.
\begin{table}[!h]
  \small
  \centering
  \renewcommand{\arraystretch}{1}
  \centering
  \caption{An example for computing $cost(\vec{B}, \vec{E})$ and $cost(s, B, E)$.}
    \begin{tabular}{c|cccc|c|cccc|c|c|c}
    \hline
         & & \multicolumn{2}{c}{$\vec{B}$} &       & $Q$     &       & \multicolumn{2}{c}{$\vec{E}$} &       & \multirow{2}[0]{*}{$cost(\vec{B}, \vec{E})$} & \multirow{2}[0]{*}{$cost(s_3, B, E)$} & \multirow{2}[0]{*}{$cost(s_4, B, E)$} \\
    \cline{2-10}
          & & $s_1$    & $s_2$    &       & $\{s_3, ~ s_4\}$ &       & $s_5$    & $s_6$    &       &       &       &  \\
    \hline
    $a_1$   & & X     & .     &       & \{X, ~ $\cdot$\}   &       & X     & X     &       & $c(a_1)(d(s_2)+ d(s_4))$ & $0$     & $0$ \\
    $a_2$   & & X     & .     &       & \{X, ~ $\cdot$\}   &       & .     & .     &       & $c(a_2)d(s_2)$ & $0$     & $c(a_2)d(s_4)$ \\
    $a_3$   & & .     & .     &       & \{X, ~ $\cdot$\}   &       & .     & X     &       & $c(a_3)d(s_5)$ & $c(a_3)d(s_4)$ & $0$ \\
    $a_4$   & & .     & X     &       & \{X, ~ $\cdot$\}   &       & .     & .     &       & $0$     & $0$     & $c(a_4)d(s_4)$ \\
    $a_5$   & & .     & .     &       & \{X, ~ $\cdot$\}   &       & .     & .     &       & $0$     & $0$     & $0$ \\
    \hline
    \end{tabular}%
  \label{tab:z}%
\end{table}%

Define $o(Q) = a(S-Q) \cap a(Q)$ as the set of actors required by scenes in both $Q$ and $S-Q$ \citep{de2011solving}. Then, $cost(s,B,E)$ can be mathematically computed by:
\begin{small}
\begin{align}
cost(s,B,E)=&d(s)\times c\big{(}o(B)-o(E)-a(s)\big{)} \nonumber \\
&+\sum_{s' \in Q - \{s\}} d(s')\times \bigg{(} c\Big{(}\big{(}a(s)-o(B)\big{)}\cap o(E)\Big{)} - c\Big{(}\big{(}a(s)-o(B)\big{)}\cap o(E)\cap a(s')\Big{)} \bigg{)}, \label{eqn:cost}
\end{align}
\end{small}
where $c(G)$ is the total daily cost of all actors in $G \subseteq A$, i.e., $c(G) = \sum_{a \in G} c(a)$.

We use Table \ref{tab:2} to explain Expression (\ref{eqn:cost}). All actors can be classified into  16 patterns according to whether they are required by the scenes in sets $B$, $\{s\}$, $Q - \{s\}$ and $E$. If an actor of some pattern is required by at least one scene in some set, the corresponding cell in columns 2 -- 5 is filled with a sign ``$\mathbb{X}$''; otherwise it is filled with a sign ``$\cdot$''. In columns 6 -- 12, if an actor of some pattern is included in some actor set, the corresponding cell is filled with ``1''; otherwise, it is filled with ``0''. For example, for patten 2 actors that has $(B, \{s\}, Q - \{s\}, E) = (\cdot, \mathbb{X}, \mathbb{X}, \mathbb{X})$, we can derive that all actors of this patten must be included in sets $o(E), a(s), a(s) - o(B)$ and $(a(s) - o(B)) \cap o(E)$ and cannot exist in sets $o(B)$, $o(B) - o(E)$ and $o(B) - o(E) - a(s)$.

From Table \ref{tab:2}, we can observe that set $o(B) - o(E) - a(s)$ only contains type 1 actors that have patten $(B, \{s\}, Q - \{s\}, E) = ( \mathbb{X}, \cdot, \mathbb{X}, \cdot)$. Thus, the first component of Expression (\ref{eqn:cost}) corresponds to type 1 actors. Set $\big{(}a(s)-o(B)\big{)}\cap o(E)$ contains type 2 actors that have either pattern $(B, \{s\}, Q - \{s\}, E) = (\cdot, \mathbb{X}, \mathbb{X}, \mathbb{X})$ or pattern $(B, \{s\}, Q - \{s\}, E) = (\cdot, \mathbb{X}, \cdot, \mathbb{X})$. The second component of Expression (\ref{eqn:cost}) is the holding cost of type 2 actors during the shooting days for the scenes in $Q - \{s\}$.

\begin{table}[!h]
  \centering
  \caption{The table for explaining Expression (\ref{eqn:cost}).}
    \scalebox{0.7}{
    \begin{tabular}{c|cccc|ccccccc}
    \hline
     Actor pattern     & $B$     & $\{s\}$   & $Q-\{s\}$ & $E$     & $o(B)$  & $o(E)$  & $a(s)$  & $o(B)-o(E)$ & $o(B)-o(E)-a(s)$ & $a(s)-o(B)$ & $(a(s)-o(B))\cap o(E)$ \\
    \hline
    $1$    & $\mathbb{X}$     & $\mathbb{X}$     & $\mathbb{X}$     & $\mathbb{X}$     & 1     & 1     & 1     & 0     & 0     & 0     & 0 \\
    $2$    & $\cdot$     &$\mathbb{X}$     & $\mathbb{X}$     & $\mathbb{X}$     & 0     & 1     & 1     & 0     & 0     & 1     & 1 \\
    $3$    & $\mathbb{X}$     & $\mathbb{X}$    & $\mathbb{X}$     & $\cdot$      & 1     & 0     & 1     & 1     & 0     & 0     & 0 \\
    $4$    & $\cdot$      & $\mathbb{X}$    & $\mathbb{X}$     & $\cdot$      & 0     & 0     & 1     & 0     & 0     & 1     & 0 \\
    $5$    & $\mathbb{X}$     & $\mathbb{X}$     & $\cdot$      & $\mathbb{X}$     & 1     & 1     & 1     & 0     & 0     & 0 & 0 \\
    $6$    & $\cdot$      & $\mathbb{X}$     & $\cdot$      & $\mathbb{X}$     & 0     & 1     & 1     & 0     & 0     & 1     & 1 \\
    $7$    & $\mathbb{X}$     & $\mathbb{X}$    & $\cdot$      & $\cdot$      & 1     & 0     & 1     & 1     & 0     & 0     & 0 \\
    $8$    & $\cdot$      & $\mathbb{X}$     & $\cdot$      & $\cdot$      & 0     & 0     & 1     & 0     & 0     & 1     & 0 \\
    ${9}$   & $\mathbb{X}$     & $\cdot$      & $\mathbb{X}$     & $\mathbb{X}$     & 1     & 1     & 0     & 0     & 0     & 0     & 0 \\
    ${10}$   & $\cdot$      & $\cdot$      & $\mathbb{X}$     & $\mathbb{X}$     & 0     & 1     & 0     & 0     & 0     & 0     & 0 \\
    ${11}$   & $\mathbb{X}$     & $\cdot$      & $\mathbb{X}$     & $\cdot$      & 1     & 0     & 0     & 1     & 1     & 0     & 0 \\
    ${12}$   & $\cdot$      & $\cdot$     & $\mathbb{X}$     & $\cdot$      & 0     & 0     & 0     & 0     & 0     & 0     & 0 \\
    ${13}$   & $\mathbb{X}$     & $\cdot$      & $\cdot$      & $\mathbb{X}$     & 1     & 1     & 0     & 0     & 0     & 0     & 0 \\
    ${14}$   & $\cdot$      & $\cdot$     & $\cdot$      & $\mathbb{X}$     & 0     & 0     & 0     & 0     & 0     & 0     & 0 \\
    ${15}$   & $\mathbb{X}$     &$\cdot$      & $\cdot$      & $\cdot$      & 0     & 0     & 0     & 0     & 0     & 0     & 0 \\
    ${16}$   &$\cdot$     & $\cdot$      &$\cdot$     & $\cdot$      & 0     & 0     & 0     & 0     & 0     & 0     & 0 \\
    \hline
    \end{tabular}%
  \label{tab:2}%
  }
\end{table}%

\subsection{Preprocessing}
\label{sec:pre}
The holding costs of all fixed actors will not change in the later stages of the search. We use set $A_N$ to contain all {\em non-fixed actors}, namely $A_N = \{a_i \in A: a_i \notin a(B) \cap a(E)\}$. When solving problem $P(\vec{B}, Q, \vec{E})$, we only need to consider the actors in $A_N$. The problem $P(\vec{B}, Q, \vec{E})$ can be further simplified as:
\begin{itemize}
\item We remove from $A_N$ all actors that are required by only one scene. This is because such actors will not bring about extra holding cost.
\item We exclude from $A_N$ all non-fixed actors that are not required by the scenes in $Q$.
\item If scenes $s_1$ and $s_2$ satisfy $a(s_1)\cap A_N = a(s_2)\cap A_N$, then we replace them with a single scene with duration $d(s) = d(s_1) + d(s_2)$ since they can be regarded as duplicate scenes. The correctness of merging duplicate scenes has been proved by \citet{de2011solving}.
\end{itemize}

The example shown in Table \ref{tab:pre} illustrates the preprocessing steps. In the problem given by Table \ref{tab:pre1}, actor $a_4$ is fixed and actor $a_5$ is not required by the scenes in $Q = \{s_1, s_2, s_3, s_4\}$. Therefore, we can remove actors $a_4$ and $a_5$ to make $A_N = \{a_1, a_2, a_3\}$. Now since $a(s_2)\cap A_N = a(s_3) \cap A_N = \{a_1, a_2, a_3\}$, we merge scenes $s_2$ and $s_3$. After these preprocessing steps, we can get a new problem as shown in Table \ref{tab:pre2}.

\begin{table}[!h]
\caption{An example to illustrate preprocessing steps.}
\label{tab:pre}
\begin{minipage}[t]{0.5\linewidth}
\subtable[Before preprocessing.]{
  \label{tab:pre1}
  \scriptsize
  \centering
    \begin{tabular}{c|c|cccc|c}
    \hline
          & \multirow{2}[0]{*}{$B$} & & \multicolumn{2}{c}{$Q$} & & \multirow{2}[0]{*}{$E$} \\
    \cline{3-6}
          &       & $s_1$ & $s_2$ & $s_3$ & $s_4$ &  \\
    \hline
    $a_1$    & $\mathbb{X}$     & X     & X     & X     &  X     & $\cdot$ \\
    $a_2$    & $\mathbb{X}$     & $\cdot$     & X     & X     &    X   & $\cdot$ \\
    $a_3$    & $\cdot$     & X     & X     & X     &  $\cdot$     & $\mathbb{X}$ \\
    $a_4$    & $\mathbb{X}$     & $\cdot$     & X     &$\cdot$     & $\cdot$     & $\mathbb{X}$ \\
    $a_5$    & $\cdot$  & $\cdot$    &$\cdot$    &$\cdot$    &  $\cdot$     & $\mathbb{X}$ \\
    \hline
    \end{tabular}%
}
\end{minipage}
\hspace{0.5cm}
\begin{minipage}[t]{0.5\linewidth}
\subtable[After preprocessing.]{
  \label{tab:pre2}
  \scriptsize
  \centering
    \begin{tabular}{c|c|ccc|c}
    \hline
          & \multirow{2}[0]{*}{$B$} & & \multicolumn{1}{c}{$Q$} & & \multirow{2}[0]{*}{$E$} \\
    \cline{3-5}
          &       & $s_1$ & $\{s_2,s_3\}$ & $s_4$ &  \\
    \hline
    $a_1$    & $\mathbb{X}$     & X     & X     &  X     & $\cdot$ \\
    $a_2$    & $\mathbb{X}$     & $\cdot$     & X     &    X   & $\cdot$ \\
    $a_3$    & $\cdot$    & X     & X     &  $\cdot$     & $\mathbb{X}$ \\
    \hline
    \end{tabular}%
}
\end{minipage}
\end{table}

\subsection{Lower Bound to Future Cost}
\label{sec:lb}
In \citet{de2011solving}, the authors proposed a lower bound to the future cost. They generated two lower bounds using ($o(B) - F$, $Q$) and  ($o(E) - F$, $Q$) as input information, and claimed that the sum of these two lower bounds is still a lower bound (denoted by $L_0$) to the future cost. The reader is encouraged to refer to \citet{de2011solving} for the details of this lower bound.

%We denote by $L_0$ the lower bound to future cost generated by the method of \citet{de2011solving}.

In this subsection, we present a new implementation of \textbf{lower}($B$,$Q$,$E$). Suppose $\sigma$ is an arbitrary permutation of the scenes in $Q$. We denote by $x_i$ the holding cost of actor $a_i$ during the period of shooting the scenes in $Q$ with the order specified by permutation $\sigma$. If \textbf{lower}($B$,$Q$,$E$) = $\min_{\sigma}\{\sum_{i\in A_N} x_i\}$, we get the minimum possible future cost. However, it is impossible to get the value of $\min_{\sigma}\{\sum_{i\in A_N} x_i\}$ unless all $\sigma$ are checked. In the following context, we describe a method for a lower bound to $\min_{\sigma}\{\sum_{i\in A_N} x_i\}$.

If an actor $a_i$ satisfies $a_i \notin a(B)$, $a_i \notin a(E)$ and $a_i \in a(Q)$, the lowest possible holding cost of this actor during the the period of shooting the scenes in $Q$ may be zero. Therefore, we only consider the actors in set $A'_N = (o(B)-F) \cup (o(E)-F) \subseteq A_N$. For any two different actors $a_i, a_j \in A'_N$, we can derive a constraint $x_i + x_j \geq c_{i,j}$, where $c_{i, j}$ is a constant computed based on the following four cases:

{\em Case 1}: $a_i, a_j \in o(B) - F$. Let $a_i(s) =$ ``X'' if actor $a_i$ is required by scene $s$ and $a_i(s) =$ ``$\cdot$'' otherwise. For any scene $s \in Q$, the tuple $(a_i(s), a_j(s))$ must have one of the following four patterns: (X, X), (X, $\cdot$), ($\cdot$, X), ($\cdot$, $\cdot$). First, we schedule all scenes with pattern (X, X) immediately after the scenes in $B$ and schedule all scenes with pattern ($\cdot, \cdot$) immediately before the scenes in $E$. Second, we group the scenes with (X, $\cdot$) and the scenes with ($\cdot$, X) into two sets. Third, we schedule these two set of scenes in the middle of the permutation, creating two schedules as shown in Table \ref{tab:lb3}. If only actors $a_i$ and $a_j$ are considered, the optimal schedule must be either one of these two schedules. The value of $c_{i, j}$ is set to the holding cost of the optimal schedule. For the schedule in Table \ref{tab:lb3}(a), if we define $S_1 = \{s \in Q|(a_i(s), a_j(s)) = (X, \cdot)\}$, then the holding cost is $c(a_j)\times d(S_1)$, where $d(S_1) = \sum_{s\in S_1}d(s)$. Similarly, for the schedule in Table \ref{tab:lb3}(b), we have a holding cost $c(a_i)\times d(S_2)$, where $S_2 = \{s \in Q|(a_i(s), a_j(s)) = (\cdot, X)\}$. Accordingly, we set $c_{i, j} = \min\{c(a_j)\times d(S_1), c(a_i)\times d(S_2)\}$.

\begin{table}[!h]
\caption{Two schedules in Case 1.}
\label{tab:lb3}
\begin{minipage}[t]{0.5\linewidth}
\subtable[The first schedule.]{
  \scriptsize
  \centering
    \begin{tabular}{c|c|cccccccc|c}
    \hline
          & \multirow{2}[0]{*}{$B$} & & \multicolumn{6}{c}{$Q$} & & \multirow{2}[0]{*}{$E$} \\
    \cline{3-10}
          &       & $s_1$ & $s_2$ & $s_3$ & $s_4$ & $s_5$ & $s_6$ & $s_7$ & $s_8$ \\
    \hline
    $a_i$    & $\mathbb{X}$     & X     & X     & X     &  X     & $\cdot$     &  $\cdot$      &  $\cdot$     &  $\cdot$     &  $\cdot$     \\
    $a_j$    & $\mathbb{X}$     & X     & X     & $\cdot$    &  $\cdot$     & X     &  X     &  $\cdot$      & $\cdot$     &  $\cdot$      \\
    \hline
    \end{tabular}%
}
\end{minipage}
\hspace{0.5cm}
\begin{minipage}[t]{0.5\linewidth}
\subtable[The second schedule.]{
  \scriptsize
  \centering
    \begin{tabular}{c|c|cccccccc|c}
    \hline
          & \multirow{2}[0]{*}{$B$} & & \multicolumn{6}{c}{$Q$} & & \multirow{2}[0]{*}{$E$} \\
    \cline{3-10}
          &       & $s_1$ & $s_2$ & $s_5$ & $s_6$ & $s_3$ & $s_4$ & $s_7$ & $s_8$ \\
    \hline
    $a_i$    & $\mathbb{X}$     & X     & X     &$\cdot$      &  $\cdot$     & X     &  X     & $\cdot$     &  $\cdot$     &  $\cdot$     \\
    $a_j$    & $\mathbb{X}$     & X     & X     & X     &  X     & $\cdot$     & $\cdot$      & $\cdot$     &  $\cdot$     &  $\cdot$      \\
    \hline
    \end{tabular}%
}
\end{minipage}
\end{table}

{\em Case 2}: $a_i, a_j \in o(E) - F$. We schedule all scenes with pattern (X, X) immediately before the scenes in $E$ and schedule all scenes with pattern ($\cdot, \cdot$) immediately after the scenes in $B$. The remaining analysis is similar to that in Case 1.

{\em Case 3}: $a_i\in o(B)-F$ and $a_j\in o(E)-F$.  We schedule all scenes with pattern (X, $\cdot$) immediately after the scenes in $B$ and schedule all scenes with pattern ($\cdot$, X) immediately before the scenes in $E$. If there does not exist a scene with pattern (X, X), the holding cost may be zero and thus $c_{i, j}$ is set to zero; otherwise $c_{i, j}$ is set to $\min\{ c(a_i), c(a_j)\} \times d(S_0)$, where $S_0 = \{ s \in Q | (a_i(s), a_j(s)) = $ ($\cdot$, $\cdot)$ \}, which can be observed from Table \ref{tab:lb2}.

\begin{table}[!h]
\caption{Two schedules in Case 3.}
\label{tab:lb2}
\begin{minipage}[t]{0.5\linewidth}
\subtable[The first schedule.]{
  \scriptsize
  \centering
    \begin{tabular}{c|c|cccccccc|c}
    \hline
          & \multirow{2}[0]{*}{$B$} & & \multicolumn{6}{c}{$Q$} & & \multirow{2}[0]{*}{$E$} \\
    \cline{3-10}
          &       & $s_1$ & $s_2$ & $s_3$ & $s_4$ & $s_5$ & $s_6$ & $s_7$ & $s_8$ \\
    \hline
    $a_i$    & $\mathbb{X}$     & X     & X     & X     &  X     & $\cdot$     & $\cdot$     & $\cdot$     &  $\cdot$    & $\cdot$    \\
    $a_j$    & $\cdot$  &$\cdot$   &$\cdot$   & X     &  X     & $\cdot$    &  $\cdot$    &  X     &  X     &  $\mathbb{X}$     \\
    \hline
    \end{tabular}%
}
\end{minipage}
\hspace{0.5cm}
\begin{minipage}[t]{0.5\linewidth}
\subtable[The second schedule.]{
  \scriptsize
  \centering
    \begin{tabular}{c|c|cccccccc|c}
    \hline
          & \multirow{2}[0]{*}{$B$} & & \multicolumn{6}{c}{$Q$} & & \multirow{2}[0]{*}{$E$} \\
    \cline{3-10}
          &       & $s_1$ & $s_2$ & $s_5$ & $s_6$ & $s_3$ & $s_4$ & $s_7$ & $s_8$ \\
    \hline
    $a_i$    & $\mathbb{X}$     & X     & X     & $\cdot$     &  $\cdot$    & X     &  X     & $\cdot$     &  $\cdot$     &  $\cdot$     \\
    $a_j$    & $\cdot$    & $\cdot$    & $\cdot$    & $\cdot$     &  $\cdot$     & X     &  X     &  X     &  X     &  $\mathbb{X}$     \\
    \hline
    \end{tabular}%
}
\end{minipage}
\end{table}

{\em Case 4}: $a_i\in o(E)-F$ and $a_j\in o(B)-F$. This case is the same as Case 3.

A valid lower bound to the future cost (i.e., the value of \textbf{lower}($B$,$Q$,$E$)) can be obtained by solving the following linear programming model:
\begin{align}
(LB)~~z^{LB}=& \min~~\sum_{a_i \in A'_N} x_i \label{eqnl:obj} \\
\mbox{s.t.}~ & x_i + x_j \geq c_{i, j}, ~
		\forall~ a_i, a_j \in A'_N, i\neq j \label{eqnl:c1} \\
	& x_i \geq 0, ~
		\forall~  a_i \in A'_N \label{eqnl:c2}
\end{align}
The value of $z^{LB}$ must be a valid lower bound to $\min_{\sigma}\{\sum_{i\in A_N} x_i\}$. If the daily holding cost of actor $a_i$ is an integral number, decision variable $x_i$ should be integer. When all variables $x_i$ are integers, the model $(LB)$ is an NP-hard problem  since it can be easily reduced to the \textit{minimum vertex cover problem} \citep{karp1972reducibility}. If all variables $x_i$ are treated as real numbers, this model can be solved by a liner programming solver. For some instances, the $(LB)$ model needs to be solved more than two million times. To save computation time, we apply the following two heuristic approaches to rapidly produce two lower bounds, i.e., $L_1$ and $L_2$, to $z^{LB}$. Obviously, $L_1$ and $L_2$ are also valid lower bounds to the future cost.

{\em Approach 1:} Sum up the left-hand-side and righ-hand-side of Equations (\ref{eqnl:c1}), generating $(|A'_N|-1)\sum_{a_i \in A'_N} x_i \geq \sum_{a_i, a_j \in A'_N, i\neq j} c_{i,j}$. The valid lower bound $L_1$ is defined as:
     \begin{align}
   L_1 =  \sum_{a_i, a_j \in A'_N, i\neq j} c_{i,j}/(|A'_N| - 1). \nonumber
     \end{align}

{\em Approach 2:} Sort $c_{i,j}$ in descending order. If we select a $c_{i, j}$, we call the corresponding $x_i$ and $x_j$ {\em marked}. Beginning from the largest $c_{i, j}$, we select all $c_{i, j}$ whose $x_i$ and $x_j$ are not marked until all $x_i$ are marked. The valid lower bound $L_2$ equals the sum of all selected $c_{i, j}$. This approach was termed the \textit{greedy matching algorithm} \citep{drake2003simple}. To demonstrate the process of computing $L_2$, we consider the following six constraints:
\begin{align}
x_1 + x_2 \geq 2,~
x_1 + x_3 \geq 7,~
x_1 + x_4 \geq 6, \nonumber\\
x_2 + x_3 \geq 12,~
x_2 + x_4 \geq 8,~
x_3 + x_4 \geq 5. \nonumber
\end{align}
We first select $c_{2, 3} = 12$ and mark $x_2$ and $x_{3}$. Then, we can only select $c_{1, 4} = 6$ since $x_1$ and $x_4$ have not been marked. Now all $x_i$ are marked and the value of $L_2$ equals 18.

In our algorithm, we set \textbf{lower}($B$, $Q$, $E$) = $\max\{L_0, L_1, L_2\}$.

\subsection{Caching Search States}
\label{sec:cac}
In \citet{de2011solving}, the talent scheduling problem was solved by a double-ended dynamic programming (DP) algorithm, where a DP state is represented by $\langle B, E\rangle$. The DP algorithm stores the best value of each examined state, denoted by $\langle B, E\rangle.value$, which equals the minimum past cost of all search paths associated with sets $B$ and $E$.

%The recursive equation can be written as:
%\begin{align}
%\langle B, E\rangle.value =  \langle E, B \rangle.value = \min_{s\in B}\big{\{}\langle E, B-\{s\}\rangle.value + cost(s, B - \{s\}, E)\big{\}} %\nonumber
%\end{align}

We embed the DP process in the branch-and-bound framework by use of {\em memoization} technique \citep{michie1968memo}. More precisely, when the search process reaches a tree node $P(\vec{B}, Q, \vec{E})$, it first checks whether the value of $cost(\vec{B}, \vec{E})$ is less than the current $\langle B, E\rangle.value$. If so, it updates $\langle B, E\rangle.value$ by $cost(\vec{B}, \vec{E})$; otherwise, the current node must be dominated by some node and therefore can be safely discarded.

A better state representation for the DP algorithm is $\langle o(B),o(E), Q \rangle$, where $Q = S - B - E$; this was discussed by \citet{de2011solving} as follows. The cost of scheduling the scenes in $Q = S - B -E$ depends on $o(B)$ and $o(E)$ rather than $B$ and $E$. Suppose $\vec{B}\vec{Q}\vec{E}$ and $\vec{B'}\vec{Q}\vec{E'}$ are two permutations of $S$, where $B$, $Q$, $E$, $B'$ and $E'$ are the corresponding sets of scenes. If $o(B) = o(B')$ and $o(E) = o(E')$, then the holding costs incurred by $\vec{Q}$ in these two permutations are equivalent. Moreover, if there are two states $\langle o(B), o(E), Q\rangle$ and $\langle o(B'), o(E'), Q\rangle$ that have $o(B) = o(E')$ and $o(E) = o(B')$, they are equivalent according to the symmetric property of the problem. Thus, we only need to memoize the state $\langle o(B),o(E),Q\rangle$ that satisfies $o(B) \leq o(E)$. We compare $o(B)$ with $o(E)$ based on the lexicographical order of the actor indices. For example, given $o(B) =\{a_1, a_2, a_4, a_5\} $ and $o(E) = \{a_1, a_3, a_6, a_7\}$, we have $o(B) \leq o(E)$ since the index of $a_2$ is less than that of $a_3$.

We also use the memoization technique to prune the search tree node. The process of checking whether a given node associated with problem $P(\vec{B}, Q, \vec{E})$ can be pruned is depicted in Algorithm \ref{alg:4}. All states are stored in a hash table {\em hashTable}. This algorithm first designates a storage slot in the hash table for state $\langle o(B),o(E), Q \rangle$ using function {\bf hash($o(B), o(E), Q$)}. If the storage slot contains the state and the current value of the state is less than or equal to $cost(\vec{B}, \vec{E})$, the algorithm returns {\em true}, implying that the given node can be pruned (see line \ref{algo2:line:3}). Next, it checks whether the state $\langle o(B), o(E), Q - \{s\} \rangle$ ($s \in Q$) exists in the hash table and has a value less than or equal to $cost(\vec{B}, \vec{E})$ (see lines \ref{algo2:line:4} -- \ref{algo2:line:7}). If such states exist, the given node can also be pruned. The correctness of this pruning condition is guaranteed by Property \ref{pro:1}, which was derived from the second theorem in \citet{de2011solving}.

\begin{small}
\begin{algorithm}[h]
\SetAlgoNoLine
\LinesNumbered
\SetCommentSty{textit}
\small
\SetKwFunction{Check}{Check}
\SetKwInput{Func}{Function} \BlankLine
\Func{$\textbf{check}(\vec{B}, Q, \vec{E}$)}
$pc = cost(\vec{B}, \vec{E})$ \;
$index: = \textbf{hash}(o(B),o(E),Q)$ \;
\lIf {$hashTable[index].state=\langle o(B),o(E),Q\rangle$ \textbf{and} $hashTable[index].value\leq pc$} {
	\Return true \label{algo2:line:3}\;
}
\ForEach {$s\in Q$} {  \label{algo2:line:4}
	$index2:=\textit{hash}(o(B),o(E),Q-\{s\})$  \;
	\lIf {$hashTable[index2].state=\langle o(B),o(E),Q-\{s\}\rangle$ \textbf{and} $hashTable[index2].value\leq pc$} {
		\Return true  \;
	}
}\label{algo2:line:7}
\If {$\textbf{replace}(index, pc)$} {
	$hashTable[index].state:= \langle o(B),o(E),Q\rangle$ \;
	$hashTable[index].value:= pc$ \;
}
\Return false \;
\caption{The process of checking whether a given search node can be pruned.}
\label{alg:4}
\end{algorithm}
\end{small}

\begin{property}
\label{pro:1}
Suppose $\vec{B}\vec{Q}\vec{E}$ and $\vec{B'}\vec{Q'}\vec{E'}$ are two permutations of $S$, where $B$, $Q$, $E$, $B'$, $Q'$ and $E'$ are the corresponding sets of scenes. If $o(B) = o(B')$, $o(E) = o(E')$, $Q \subseteq Q'$ and the scenes in $\vec{Q}$ follow the order in which they appear in $\vec{Q'}$, then the holding cost incurred by $\vec{Q}$ is not greater than that incurred by $\vec{Q'}$.
\end{property}

Ideally, the hash function should assign each state to a unique storage slot, i.e., no hash collisions happen. However, this ideal situation is rarely achievable due to the huge number of states and the inadequate storage space. When solving the talent scheduling problem, we do not have sufficient storage space to store the exponential number of search states and therefore different states may be assigned by the hash function to the same storage slot, leading to hash collisions. To resolve this issue, we employ a mechanism called {\em direct mapped caching scheme}. Assume the direct mapped cache consists of $C$ slots, each of which can only store one item. If an item is to be stored in a slot that already contains another item (i.e., a hash collision occurs), it may either replace the existing item or be discarded, which is decided by function \textbf{replace}($index$, $pc$). Several previous articles, such as \citet{hilden1976elimination} and \citet{pugh1988improved}, have discussed the replacement strategies implemented in \textbf{replace}($index$, $pc$). In this work, we tried {\em latest} and {\em greedy} caching strategies. The first strategy deals with the hash collisions by simply overwriting the cache slot while the second one stores in the cache slot the item that has smaller value.

The direct mapped caching scheme can effectively prune the search nodes using limited storage space. When a state is revisited again but it has been removed from the cache during the previous stages, the search can still continue to explore its corresponding subtree. In Section \ref{sec:exp}, we experimentally analyze the impact of different values of $C$ and the two replacement strategies on the performance of our branch-and-bound algorithm.

\subsection{Dominance Rules}
Dominance rules are widely used in branch-and-bound algorithms \citep{Zhang2012Single,Braune2012Exact,Ranjbar2012Two,Kellegoz2012Efficient} and dynamic programming algorithms \citep{dumas1995optimal,mingozzi1997dynamic,Rong2013Reduction} for reducing search space. The purpose of dominance rules is to determine when the partial solution represented by a node in the search tree is dominated by another node; if so, the node need not be further explored and can be safely pruned. In our branch-and-bound algorithm, two dominance rules are employed to reduce the search space.

\subsubsection{Dominance Rule 1}
At a branch-and-bound tree node associated with problem $P(\vec{B}, Q, \vec{E})$, we suppose that scene $s_1$ is the scene to be scheduled immediately after $B$ and scene $s_2$ belongs to $Q - \{s_1\}$. If $a(s_1)\cup o(B)\supseteq a(s_2) \cup o(B)$ and $a(s_1) \cup o(E)\subseteq a(s_2) \cup o(E)$, then the branch associated with scene $s_1$ can be ignored.

Tables \ref{tab:er11} -- \ref{tab:er12} are used to explain this dominance rule. In Table \ref{tab:er11}, $Q=\{s_1, s_2\}\cup \Omega_1 \cup \Omega_2$, where $\Omega_1$ and $\Omega_2$ are two arbitrary subsets of $Q - \{s_1, s_2\}$ and $\Omega_1 \cap \Omega_2 = \emptyset$. Actors in $A_N$ can be classified into twelve patterns according to whether they are required by the scenes in sets $B$, $E$, $\{s_1\}$ and $\{s_2\}$. Since we do not need the information related to $\Omega_1$ and $\Omega_2$, all cells in columns 4 and 6 remain empty. Similar to Table \ref{tab:2}, the numbers $1$ and $0$ in the right part of Table \ref{tab:er11} indicate whether an actor of some pattern is included in the corresponding actor set.

In the absence of the information in columns 4 and 6, we cannot directly judge whether patten 4 actors are included in $o(B)$ and whether patten 8 actors are included in $o(E)$. However, we know that all remaining actors are non-fixed and must be required by the scenes in $Q$. In other words, if some pattern 4 and 8 actors are kept in $A_N$, then they must be required by some scene in $\Omega_1 \cup \Omega_2 $. Therefore, we fill the corresponding cells with ``1'' (see the numbers in bold in Table \ref{tab:er11}).

We list in the left part of Table \ref{tab:er12} all actor patterns that satisfy the conditions $a(s_1)\cup o(B)\supseteq a(s_2)\cup o(B)$ and $a(s_1)\cup o(E)\subseteq a(s_2)\cup o(E)$. Table \ref{tab:er12} shows that branching to scene $s_1$ is dominated by branching to scene $s_2$. After exchanging the positions of scenes $s_1$ and $s_2$, the holding costs for pattern 1, 4 -- 5, 8 -- 9 and 12 actors remain unchanged while the holding costs for pattern 3 and 6 actors are probably reduced. Thus, scheduling scene $s_2$ immediately after $B$ must result in less or equal holding cost than scheduling scene $s_1$ at that position.

\begin{table}[!h]
  \centering
  \addtolength{\tabcolsep}{-1pt}
  \caption{The table for explaining dominance rules.}
  \scalebox{0.65}{
    \begin{tabular}{c|cccccc|ccccccc}
    \hline
    \multirow{2}[0]{*}{Actor pattern} & \multirow{2}[0]{*}{$B$} & \multirow{2}[0]{*}{$\{s_1\}$} & \multirow{2}[0]{*}{$\Omega_2$} & \multirow{2}[0]{*}{$\{s_2\}$} & \multirow{2}[0]{*}{$\Omega_3$} & \multirow{2}[0]{*}{$E$} & \multirow{2}[0]{*}{$o(B)$} & \multirow{2}[0]{*}{$o(E)$} & \multirow{2}[0]{*}{$a(s_1)\cup o(E)$} & \multirow{2}[0]{*}{$a(s_1)\cup o(B)$} & \multirow{2}[0]{*}{$a(s_2)\cup o(E)$} & \multirow{2}[0]{*}{$a(s_2)\cup o(B)$} & $(a(s_1)\cup o(B))\cap$\\
          &       &       &       &       &       &       &       &       &       &       &       &       & $(a(s_2)\cup o(E))$\\
    \hline
   % $1$    & $\mathbb{X}$     & $\mathbb{X}$     &       & $\mathbb{X}$     &      & $\mathbb{X}$      & 1     & 1     & 1     & 1     & 1     & 1     & 1     \\
   % $2$    & $\mathbb{X}$     & $\mathbb{X}$     &       & $\cdot$     &      & $\mathbb{X}$     & 1     & 1     & 1     & 1     & 1     & 1     & 1     \\
   % $3$    & $\mathbb{X}$     & $\cdot$     &       & $\mathbb{X}$     &      & $\mathbb{X}$      & 1     & 1     & 1     & 1     & 1     & 1     & 1     \\
   % $4$    & $\mathbb{X}$     & $\cdot$     &       & $\cdot$     &      & $\mathbb{X}$      & 1     & 1     & 1     & 1     & 1     & 1     & 1     \\
    $1$    & $\mathbb{X}$     & $\mathbb{X}$     &       & $\mathbb{X}$     &      & $\cdot$      & 1     & 0     & 1     & 1     & 1     & 1     & 1     \\
    $2$    & $\mathbb{X}$     & $\mathbb{X}$     &       & $\cdot$     &      & $\cdot$      & 1     & 0     & 1     & 1     & 0     & 1     & 0     \\
    $3$    & $\mathbb{X}$     & $\cdot$     &       & $\mathbb{X}$     &      & $\cdot$      & 1     & 0     & 0     & 1     & 1     & 1     & 1     \\
    $4$    & $\mathbb{X}$     & $\cdot$     &       & $\cdot$     &      & $\cdot$      & {\bf 1}     & 0     & 0     & 1     & 0     & 1     & 0     \\
    $5$    & $\cdot$   &  $\mathbb{X}$     &       & $\mathbb{X}$     &      & $\mathbb{X}$      & 0     & 1     & 1     & 1     & 1     & 1     & 1     \\
    ${6}$   & $\cdot$  &  $\mathbb{X}$     &       & $\cdot$    &      & $\mathbb{X}$      & 0     & 1     & 1     & 1     & 1     & 0     & 1     \\
    ${7}$   & $\cdot$  &  $\cdot$     &       & $\mathbb{X}$    &      & $\mathbb{X}$      & 0     & 1     & 1     & 0     & 1     & 1     & 0     \\
    ${8}$   & $\cdot$  &  $\cdot$     &       & $\cdot$    &      & $\mathbb{X}$      & 0     & {\bf 1}     & 1     & 0     & 1     & 0     & 0     \\
    ${9}$   & $\cdot$ &  $\mathbb{X}$     &       & $\mathbb{X}$    &      & $\cdot$      & 0     & 0     & 1     & 1     & 1     & 1     & 1     \\
    ${10}$   & $\cdot$ &  $\mathbb{X}$     &       & $\cdot$    &      & $\cdot$      & 0     & 0     & 1     & 1     & 0     & 0     & 0     \\
    ${11}$   & $\cdot$  & $\cdot$     &       & $\mathbb{X}$    &      & $\cdot$      & 0     & 0     & 0     & 0     & 1     & 1     & 0     \\
    ${12}$   & $\cdot$  & $\cdot$     &       & $\cdot$    &      & $\cdot$      & 0     & 0     & 0     & 0     & 0     & 0     & 0     \\
    \hline
    \end{tabular}%
    }
  \label{tab:er11}%
\end{table}%

\begin{table}[!h]
  \scriptsize
  \centering
  \caption{Actor patterns before and after exchanging scenes $s_1$ and $s_2$.}
    \begin{tabular}{c|ccccccc||c|cccccc}
    \hline
      Actor pattern    & $B$     & $\{s_1\}$  & $\Omega_1$    & $\{s_2\}$  & $\Omega_2$    & $E$     &       & Actor pattern & $B$     & $\{s_2\}$  & $\Omega_1$    & $\{s_1\}$  & $\Omega_2$    & $E$ \\
    \hline
    %$1$    & $\mathbb{X}$     & $\mathbb{X}$     &       &$\mathbb{X}$     &       & $\mathbb{X}$     &       & $1$    & $\mathbb{X}$     & $\mathbb{X}$     &       & $\mathbb{X}$     &       & $\mathbb{X}$ \\
    %$2$    & $\mathbb{X}$     & $\mathbb{X}$     &       & $\cdot$     &       & $\mathbb{X}$     &       & $2$    & $\mathbb{X}$     & $\cdot$     &       & $\mathbb{X}$     &       & $\mathbb{X}$ \\
    %$3$    & $\mathbb{X}$     & $\cdot$     &       & $\mathbb{X}$     &       & $\mathbb{X}$     &       & $3$    & $\mathbb{X}$     & $\mathbb{X}$     &       & $\cdot$     &       & $\mathbb{X}$ \\
    %$4$    & $\mathbb{X}$     & $\cdot$     &       & $\cdot$     &       & $\mathbb{X}$     &       & $4$    & $\mathbb{X}$     & $\cdot$     &       & $\cdot$     &       & $\mathbb{X}$ \\
    $1$    & $\mathbb{X}$     & $\mathbb{X}$     &       & $\mathbb{X}$     &       & $\cdot$     &       & $1$    & $\mathbb{X}$     & $\mathbb{X}$     &       & $\mathbb{X}$     &       & $\cdot$ \\
    \underline{$3$}    & $\mathbb{X}$     & $\cdot$     &       & $\mathbb{X}$     &       & $\cdot$     &       & \underline{$3$}    & $\mathbb{X}$     & $\mathbb{X}$     &       & $\cdot$     &       & $\cdot$ \\
    $4$    & $\mathbb{X}$     & $\cdot$     &       & $\cdot$     &       & $\cdot$     &       & $4$    & $\mathbb{X}$     & $\cdot$     &       & $\cdot$     &       & $\cdot$ \\
    $5$    &$\cdot$     & $\mathbb{X}$     &       & $\mathbb{X}$     &       & $\mathbb{X}$     &       & $5$    & $\cdot$     & $\mathbb{X}$     &       & $\mathbb{X}$     &       & $\mathbb{X}$ \\
    \underline{${6}$}   & $\cdot$     & $\mathbb{X}$     &       & $\cdot$     &       & $\mathbb{X}$     &       & \underline{${6}$}   &$\cdot$     & $\cdot$     &       & $\mathbb{X}$     &       & $\mathbb{X}$ \\
    ${8}$   & $\cdot$     & $\cdot$     &       & $\cdot$     &       & $\mathbb{X}$     &       & ${8}$   & $\cdot$     & $\cdot$     &       & $\cdot$    &       & $\mathbb{X}$ \\
    ${9}$   & $\cdot$     & $\mathbb{X}$     &       & $\mathbb{X}$     &       & $\cdot$     &       & ${9}$   & $\cdot$     & $\mathbb{X}$     &       & $\mathbb{X}$     &       & $\cdot$ \\
    ${12}$   & $\cdot$     & $\cdot$     &       &$\cdot$     &       & $\cdot$     &       & ${12}$   &$\cdot$     & $\cdot$     &       &$\cdot$     &       & $\cdot$ \\
    \hline
    \end{tabular}%
  \label{tab:er12}%
\end{table}%

\subsubsection{Dominance Rule 2}
At a branch-and-bound tree node associated with problem $P(\vec{B}, Q, \vec{E})$, we suppose that $s_1$ is the scene to be scheduled immediately after $B$ and $s_2$ belongs to $Q - \{s_1\}$. If $a(s_1)\cup o(B)\supseteq a(s_2)\cup o(B)$ and $c((a(s_1)\cup o(B))\cap (a(s_2)\cup o(E)))-c(a(s_2)\cup o(B))>0$, then the branch associated with scene $s_1$ can be ignored.

We list in the left part of Table \ref{tab:er13} all actor patterns that satisfy the conditions $a(s_1)\cup o(B)\supseteq a(s_2)\cup o(B)$. The right part of Table \ref{tab:er13} is the result of shifting scene $s_2$ immediately before scene $s_1$ and immediately after $B$. From Table \ref{tab:er13}, we can get the following four observations: (1) the holding costs for pattern 9 actors remain unchanged; (2) the holding costs for pattern 1, 3, 8 -- 10 and 12 actors are probably reduced; (3) the holding cost of each actor $a_i$ with pattern 2 or 4 is probably increased by $c(a_i)d(s_2)$; (4) the holding cost of each patten 6 actor $a_j$ is definitely decreased by $c(a_j)d(s_2)$. If the decreased amount (related to patten 6 actors) is greater than the increased amount (related to pattern 2 and 4 actors), then shifting scene $s_2$ immediately before scene $s_1$ must lead to a cost reduction. Given that $a(s_1)\cup o(B)\supseteq a(s_2)\cup o(B)$ is satisfied, the set $a(s_1)\cup o(B))\cap (a(s_2)\cup o(E)$ includes pattern  1, 3, 5 -- 6 and 9 actors and the set $a(s_2)\cup o(B)$ includes patterns 1 -- 5, and 9 actors. Thus, if $a(s_1)\cup o(B)\supseteq a(s_2)\cup o(B)$ and $c((a(s_1)\cup o(B))\cap (a(s_2)\cup o(E)))-c(a(s_2)\cup o(B))>0$, scheduling scene $s_2$ immediately after $B$ must result in less or equal holding cost than scheduling scene $s_1$ at that position.

\begin{table}[!h]
  \scriptsize
  \centering
  \caption{Actor patterns before and after shifting scene $s_2$ immediately before scene $s_1$.}
    \begin{tabular}{c|cccccc||c|cccccc}
    \hline
     Actor pattern     & $B$     & $\{s_1\}$  & $\Omega_1$  & $\{s_2\}$    & $\Omega_2$    & $E$     &  Actor pattern     & $B$     & $\{s_2\}$  & $\{s_1\}$  & $\Omega_1$    & $\Omega_2$    & $E$ \\
    \hline
   % $1$    & $\mathbb{X}$     & $\mathbb{X}$     &      & $\mathbb{X}$      &       & $\mathbb{X}$     & $1$    & $\mathbb{X}$     & $\mathbb{X}$     & $\mathbb{X}$     &       &       & $\mathbb{X}$ \\
   % $2$    & $\mathbb{X}$     & $\mathbb{X}$     &      & $\cdot$      &       & $\mathbb{X}$     & $2$    & $\mathbb{X}$     & $\cdot$     & $\mathbb{X}$     &       &       & $\mathbb{X}$ \\
   % $3$    & $\mathbb{X}$     & $\cdot$     &      & $\mathbb{X}$      &       & $\mathbb{X}$     & $3$    & $\mathbb{X}$     & $\mathbb{X}$     & $\cdot$     &       &       & $\mathbb{X}$ \\
   % $4$    & $\mathbb{X}$     & $\cdot$     &      & $\cdot$      &       & $\mathbb{X}$     & $4$    & $\mathbb{X}$     &$\cdot$     & $\cdot$     &       &       & $\mathbb{X}$ \\
    $1$    & $\mathbb{X}$     & $\mathbb{X}$     &      & $\mathbb{X}$      &       & $\cdot$     & $1$    & $\mathbb{X}$     & $\mathbb{X}$     & $\mathbb{X}$     &       &       & $\cdot$ \\
    \underline{$2$}    & $\mathbb{X}$     & $\mathbb{X}$     &      & $\cdot$      &       &$\cdot$     & \underline{$2$}    & $\mathbb{X}$     & $\cdot$     & $\mathbb{X}$     &       &       & $\cdot$ \\
    $3$    & $\mathbb{X}$     & $\cdot$     &      & $\mathbb{X}$      &       &$\cdot$     & $3$    & $\mathbb{X}$     & $\mathbb{X}$     &$\cdot$     &       &       & $\cdot$ \\
    \underline{$4$}    & $\mathbb{X}$     & $\cdot$     &      & $\cdot$      &       & $\cdot$     & \underline{$4$}    & $\mathbb{X}$     & $\cdot$     & $\cdot$     &       &       &$\cdot$ \\
    $5$    & $\cdot$     & $\mathbb{X}$     &      & $\mathbb{X}$      &       & $\mathbb{X}$     & $5$    & $\cdot$     & $\mathbb{X}$     & $\mathbb{X}$     &       &       & $\mathbb{X}$ \\
    \underline{${6}$}   & $\cdot$     & $\mathbb{X}$     &      & $\cdot$      &       & $\mathbb{X}$     & \underline{${6}$}   & $\cdot$     & $\cdot$     & $\mathbb{X}$     &       &       & $\mathbb{X}$ \\
    ${8}$   & $\cdot$     & $\cdot$     &      & $\cdot$      &       & $\mathbb{X}$     & ${8}$   & $\cdot$     & $\cdot$     & $\cdot$     &       &       & $\mathbb{X}$ \\
    ${9}$   & $\cdot$     & $\mathbb{X}$     &      & $\mathbb{X}$      &       &$\cdot$     & ${9}$   & $\cdot$     & $\mathbb{X}$     & $\mathbb{X}$     &       &       & $\cdot$ \\
    ${10}$   &$\cdot$     & $\mathbb{X}$     &      & $\cdot$      &       & $\cdot$     & ${10}$   & $\cdot$     & $\cdot$     & $\mathbb{X}$     &       &       & $\cdot$ \\
    ${12}$   & $\cdot$     & $\cdot$     &      & $\cdot$      &       & $\cdot$     & ${12}$   & $\cdot$     & $\cdot$     & $\cdot$     &       &       & $\cdot$ \\
    \hline
    \end{tabular}%
  \label{tab:er13}%
\end{table}%

\subsection{The Enhanced Branch-and-bound Algorithm}
\label{sub:eba}
Our enhanced branch-and-bound algorithm for the talent scheduling problem is given by Algorithm \ref{alg:3}, where the value of past cost $z$ is initialized to zero at the root node. The preprocessing stage is realized by function $\textbf{preprocess}(Q, A_N)$ (see line \ref{algo:bb:8}, Algorithm \ref{alg:3}). The state caching technique is adopted through function $\textbf{check}(\vec{B}, Q, \vec{E})$ (see line \ref{algo:bb:9}, Algorithm \ref{alg:3}). The function \textbf{isDominated}$(\vec{B}, Q, \vec{E}, A_N, z, s)$ employs the proposed two dominance rules to check whether branching to some scene $s$ is dominated by other branches. The function \textbf{lower}$(B\cup\{s\}, Q-\{s\}, E)$ returns a valid lower bound to the future cost of the problem at some search node.

\begin{algorithm}[h]
\SetAlgoNoLine
\LinesNumbered
\SetCommentSty{textit}
\small
\SetKwFunction{search}{search}
\SetKwInput{Func}{Function} \BlankLine
\Func{$\textbf{search}(\vec{B}, Q, \vec{E}, A_N$, $z$)}
\If {$Q=\emptyset$} {
	\If {$z<UB$} {
		$UB:=z$\;
		$best\_solution:=\vec{B} \circ \vec{E}$\;
	}
	\Return\;
}
$(Q, A_N):=\textbf{preprocess}(Q, A_N)$\; \label{algo:bb:8}
\lIf {$\textbf{check}(\vec{B}, Q, \vec{E})$} { \label{algo:bb:9}
	\Return\;
}
\ForEach {$s\in Q$} {
	\lIf {\textbf{isDominated}$(\vec{B}, Q, \vec{E}, A_N, z, s)$} continue\;  \label{algo:bb:11}
	$LB:= z + cost(s,B,E) + \textbf{lower}(B\cup\{s\}, Q-\{s\}, E)$\; \label{algo:bb:12}
	\lIf {$LB\geq UB$} continue\;
	$\textbf{search}(\vec{E},Q-\{s\}, \vec{B}\circ\{s\}, A_N, z + cost(s,B,E))$\;
}
\caption{The enhanced double-ended branch-and-bound algorithm for the talent scheduling problem.}
\label{alg:3}
\end{algorithm}

\section{Computational Experiments}
\label{sec:exp}
Our algorithm was coded in C++ and compiled using the g++ compiler. All experiments were run on a Linux server equipped with an Intel Xeon E5430 CPU clocked at 2.66 GHz and 8 GB RAM. The algorithm only has two parameters, namely the number ($C$) of cached states and the caching strategy used. After some preliminary experiments, we set $C = 2^{25}$ and chose the \textit{greedy} caching strategy when solving the benchmark instances. In this section, we first present our results for the benchmark instances and then compare them with the results obtained by the best two existing approaches. Finally, we exhibit by experiments the impacts of the parameters on the overall performance of the algorithm. All computation times reported here are in CPU seconds on this server. All instances and detailed results are available in the online supplement to this paper at: \url{www.computational-logistics.org/orlib/tsp}.

\subsection{Results for Benchmark Instances}
\label{sec:res}
In order to evaluate our algorithm, we conducted experiments using two benchmark data sets (Types 1 and 2), downloaded from \url{http://ww2.cs.mu.oz.au/~pjs/talent/}. The Type 1 data set was introduced by \citet{cheng1993optimal} and \citet{smith2005caching}, including seven instances, namely \textit{MobStory}, \textit{film103}, \textit{film105}, \textit{film114}, \textit{film117}, \textit{film118} and \textit{film119}. Since these instances have small sizes, ranging from $18 \times 8$ (18 scenes by 8 actors) to $28 \times 8$, they were easily solved to optimality. Table \ref{tab:r1} shows the results obtained by our branch-and-bound algorithm, the constraint programming approach in \citet{smith2005caching} and the dynamic programming algorithm in \citet{de2011solving}. From this table, we can see that our algorithm reduced the number of subproblems significantly for each instance with much less computational efforts. In our branch-and-bound algorithm, a subproblem corresponds to a search tree node. Note that the results taken from \citet{smith2005caching} were produced on a PC with 1.7 GHz Pentium M processor, and the results from \citet{de2011solving} were produced on a machine with Xeon Pro 2.4 GHz processors and 2 GB RAM.

\begin{table}[!h]
  \centering
  \addtolength{\tabcolsep}{-2pt}
  \centering
  \caption{Computational results for Type 1 Data Set.}
   \scalebox{0.65}{
    \begin{tabular}{cccccccccccccccc}
    \hline
          \multirow{2}[0]{*}{Instance}  &    & \multirow{2}[0]{*}{$m~$} & \multirow{2}[0]{*}{$n$} &       & \multicolumn{2}{c}{\citet{smith2005caching}} &       & \multicolumn{2}{c}{\citet{de2011solving}} &       & \multicolumn{2}{c}{Enhanced branch-and-bound} &       & \multirow{2}[0]{*}{Total cost} & \multirow{2}[0]{*}{Holding cost}  \\
    \cline{6-7} \cline{9-10} \cline{12-13}
          &       &       &       &       & Time (s)  & Subproblems &       & Time (s)  & Subproblems &       & Time (s)  & Subproblems &       &       &  \\
    \hline
    {MobStory} &       & 8     & 28    &       & 64.71 & 136,765 &       & 0.11  & 6,605  &       & 0.05  & 849   &       & 871   & 146 \\
    {film103} &       & 8     & 19    &       & 76.69 & 180,133 &       & 0.06  & 4,103  &       & 0.02   & 828   &       & 1031  & 187 \\
    {film105} &       & 8     & 18    &       & 16.07 & 40,511 &       & 0.02  & 1,108  &       & 0.02   & 215   &       & 849   & 110 \\
    {film114} &       & 8     & 19    &       & 127   & 267,526 &       & 0.08  & 4,957  &       & 0.03  & 2,027  &       & 867   & 143 \\
    {film116} &       & 8     & 19    &       & 125.8 & 225,314 &       & 0.16  & 13,576 &       & 0.03  & 1,937  &       & 541   & 110 \\
    {film117} &       & 8     & 19    &       & 76.86 & 174,100 &       & 0.10  & 7,227  &       & 0.02   & 987   &       & 913   & 197 \\
    {film118} &       & 8     & 19    &       & 93.1  & 205,190 &       & 0.04  & 1,980  &       & 0.02   & 537   &       & 853   & 156 \\
    {film119} &       & 8     & 18    &       & 70.8  & 144,226 &       & 0.08  & 7,105  &       & 0.02   & 580   &       & 790   & 159 \\
    \hline
    \end{tabular}%
    }
  \label{tab:r1}%
\end{table}%

The Type 2 data set was provided by \citet{de2011solving}. Following a manner almost identical to that used by \citet{cheng1993optimal}, \citet{de2011solving} randomly generated 100 instances for each combination of $n \in \{16, 18, 20, \ldots, 64\}$ and $m \in \{8, 10, 12, \ldots, 22\}$, for a total of 200 instance groups and 20,000 instances. They tried to solve these instances using their dynamic programming algorithm with a memory bound of 2GB. For each instance, if the execution did not run out of memory, they recorded the running time and the number of subproblems generated. They reported the average running time and the average number of subproblems for each Type 2 instance group with more than 80 optimally solved instances; these two average values were computed based on the solved instances.

We tried to solve all Type 2 instances using our branch-and-bound algorithm with a time limit of 10 minutes and a memory of 2GB. Our algorithm requires some memory to store the information of the search tree and a limited number of states. The amount of memory available can fully satisfy this requirement and thus the out-of-memory exception did not occur. Table \ref{tab:r2} gives the number of instances optimally solved in each Type 2 instance group, where an underline sign (``\underline{~~}'') is added to the cell associated with the instance group with less than 80 optimally solved instances. For an instance group, if our algorithm optimally solved 80 or more instances while the dynamic programming algorithm failed to achieve so, the number in its corresponding cell is marked with an asterisk ($^*$). From this table, we can see that our algorithm managed to optimally solve all instances with the number of scenes ($n$) not greater than 32 or the number of actors $(m)$ not greater than 10. However, the dynamic programming algorithm by \citet{de2011solving} only optimally solved more than 80 out of 100 instances for the instance groups with $n \leq 26$. Their approach even did not optimally solve all instances with $m = 8$ and $n = 64$. In this table, 89 out of 200 instance groups are marked with asterisks, which clearly indicates that more Type 2 benchmark instances were successfully solved to optimality by our branch-and-bound algorithm. Although our machine is more powerful, this cannot account for the dramatic difference in the number of optimally solved instances; it is reasonable to conclude that our branch-and-bound algorithm is more efficient than the dynamic programming algorithm.

\begin{table}[!h]
  \scriptsize
  \centering
  \caption{The number of optimally solved instances in each Type 2 instance group.}
    \begin{tabular}{rrrrrrrrr}
    \hline
   \multirow{2}[0]{*}{$n$} & \multicolumn{8}{c}{$m$} \\
    \cline{2-9}
    & 8~~     & 10~~    & 12~~    & 14~~    & 16~~    & 18~~    & 20~~    & 22~~ \\
    \hline
    16    & 100~   & 100~   & 100~   & 100~   & 100~   & 100~   & 100~   & 100~ \\
    18    & 100~   & 100~   & 100~   & 100~   & 100~   & 100~   & 100~   & 100~ \\
    20    & 100~   & 100~   & 100~   & 100~   & 100~   & 100~   & 100~   & 100~ \\
    22    & 100~   & 100~   & 100~   & 100~   & 100~   & 100~   & 100~   & 100~ \\
    24    & 100~   & 100~   & 100~   & 100~   & 100~   & 100~   & 100~   & 100~ \\
    26    & 100~   & 100~   & 100~   & 100~   & 100~   & 100~   & 100~   & 100~ \\
    28    & 100~   & 100~   & 100~   & 100~   & 100~   & 100~   & 100~   & $100^*$ \\
    30    & 100~   & 100~   & 100~   & 100~   & 100~   & $100^*$   & $100^*$   & $100^*$ \\
    32    & 100~   & 100~   & 100~   & 100~   & $100^*$  & $100^*$   & $100^*$   & $100^*$ \\
    34    & 100~   & 100~   & 100~   & $100^*$   & $100^*$  & $100^*$   & $100^*$   & $99^*$~ \\
    36    & 100~   & 100~   & 100~   & $100^*$    & $100^*$    & $100^*$    & $99^*$~    & $98^*$~ \\
    38    & 100~   & 100~   & $100^*$   & $100^*$   & $100^*$   & $99*$~    & $99^*$~    & $98^*$~ \\
    40    & 100~   & 100~   & $100^*$   & $99^*$~     & $100^*$  & $98^*$~    & $97^*$~    & $98^*$~  \\
    42    & 100~   & 100~   & $100^*$   & $100^*$  & $100^*$   & $96^*$~    & $97^*$~    & $94^*$~ \\
    44    & 100~   & $100^*$   & $100^*$   & $100^*$   & $97^*$~    & $99^*$~    & $93^*$~    & \underline{79}~~ \\
    46    & 100~   & $100^*$   & $100^*$   & $100^*$   & $97^*$~    & $97^*$~    & $88^*$~    & \underline{71}~~ \\
    48    & 100~   & $100^*$   & $100^*$   & $100^*$   & $99^*$~    & $96^*$~    & $88^*$~    & \underline{64}~~ \\
    50    & 100~   & $100^*$   & $100^*$   & $99^*$~    & $96^*$~    & $94^*$~    & $84^*$~    & \underline{59}~~ \\
    52    & 100~   & $100^*$   & $100^*$    & $98^*$~    & $99^*$~    & $87^*$~    & \underline{70}~~    & \underline{52}~~ \\
    54    & 100~   & $100^*$    & $98^*$~    & $96^*$~    & $89^*$~    & \underline{75}~~    & \underline{63}~~    & \underline{41}~~ \\
    56    & 100~   & $100^*$    & $99^*$~    & $98^*$~    & $93^*$~    & $85^*$~    & \underline{61}~~    & \underline{44}~~ \\
    58    & 100~   & $100^*$   & $100^*$    & $97^*$~    & $82^*$~    & \underline{77}~~    & \underline{54}~~    & \underline{37}~~ \\
    60    & 100~   & $100^*$    & $99^*$~    & $89^*$~    & $87^*$~    & \underline{72}~~    & \underline{53}~~    & \underline{34}~~ \\
    62    & 100~   & $100^*$    & $96^*$~    & $96^*$~    & \underline{75}~~    & \underline{62}~~    & \underline{50}~~    & \underline{43}~~ \\
    64    & $100^*$   & $100^*$   & $98^*$~    & $95^*$~    & \underline{74}~~    & \underline{51}~~    & \underline{43}~~    & \underline{23}~~ \\
    \hline
    \end{tabular}%
  \label{tab:r2}%
\end{table}%

Tables \ref{tab:r3} -- \ref{tab:r4} show the average running time and the average number of search nodes, respectively, over all optimally solved instances for each instance group. Like in Table \ref{tab:r2}, the instance groups with less than 80 optimally solved instances are marked with ``\underline{~~}''. From Table \ref{tab:r4}, we can easily find that the average number of search nodes generated for each instance group with ``\underline{~~}'' exceeds 3,000,000.

\begin{table}[!h]
  \scriptsize
  \centering
    \caption{The average running time (in seconds) for each Type 2 instance group.}
    \begin{tabular}{rrrrrrrrr}
    \hline
    \multirow{2}[0]{*}{$n$} & \multicolumn{8}{c}{$m$} \\
    \cline{2-9}
     & 8     & 10    & 12    & 14    & 16    & 18    & 20    & 22 \\
    \hline
    16    & 0.004 & 0.003 & 0.006 & 0.012 & 0.083 & 0.089 & 0.121 & 0.175 \\
    18    & 0.006 & 0.009 & 0.011 & 0.020 & 0.123 & 0.119 & 0.134 & 0.242 \\
    20    & 0.013 & 0.015 & 0.020 & 0.032 & 0.153 & 0.126 & 0.166 & 0.291 \\
    22    & 0.018 & 0.021 & 0.030 & 0.048 & 0.142 & 0.199 & 0.236 & 0.388 \\
    24    & 0.022 & 0.033 & 0.047 & 0.065 & 0.185 & 0.232 & 0.426 & 0.704 \\
    26    & 0.032 & 0.043 & 0.068 & 0.101 & 0.274 & 0.504 & 0.738 & 1.058 \\
    28    & 0.038 & 0.068 & 0.146 & 0.235 & 0.398 & 0.663 & 1.461 & 2.433 \\
    30    & 0.055 & 0.083 & 0.155 & 0.333 & 0.640 & 1.548 & 2.183 & 5.541 \\
    32    & 0.069 & 0.097 & 0.219 & 0.566 & 1.613 & 2.496 & 4.822 & 16.342 \\
    34    & 0.085 & 0.160 & 0.281 & 1.063 & 1.672 & 4.933 & 10.761 & 17.891 \\
    36    & 0.107 & 0.202 & 0.898 & 2.254 & 3.269 & 7.848 & 24.991 & 29.444 \\
    38    & 0.108 & 0.401 & 0.619 & 3.153 & 5.842 & 18.746 & 36.231 & 48.581 \\
    40    & 0.154 & 0.374 & 0.915 & 2.197 & 12.178 & 12.270 & 38.954 & 53.126 \\
    42    & 0.198 & 0.490 & 1.323 & 4.602 & 14.271 & 31.411 & 46.482 & 100.084 \\
    44    & 0.825 & 0.826 & 2.525 & 7.661 & 18.215 & 57.402 & 64.493 & \underline{85.409} \\
    46    & 0.935 & 2.986 & 2.604 & 9.393 & 19.240 & 35.505 & 84.877 & \underline{104.940} \\
    48    & 0.892 & 1.404 & 6.162 & 11.908 & 39.599 & 61.703 & 90.863 & \underline{107.725} \\
    50    & 0.953 & 1.791 & 10.579 & 20.588 & 48.641 & 71.460 & 111.824 & \underline{120.828} \\
    52    & 0.995 & 3.819 & 13.127 & 37.832 & 40.237 & 97.423 & \underline{108.915} & \underline{130.559} \\
    54    & 1.525 & 3.380 & 19.269 & 22.891 & 67.856 & \underline{97.367} & \underline{134.863} & \underline{163.487} \\
    56    & 1.393 & 3.770 & 15.025 & 49.464 & 50.699 & 100.411 & \underline{134.661} & \underline{149.144} \\
    58    & 1.867 & 6.014 & 30.280 & 37.760 & 61.514 & \underline{117.141} & \underline{158.239} & \underline{173.229} \\
    60    & 2.156 & 7.378 & 22.250 & 70.460 & 97.438 & \underline{146.204} & \underline{105.785} & \underline{207.466} \\
    62    & 1.626 & 10.162 & 27.478 & 64.337 & \underline{128.221} & \underline{158.739} & \underline{167.173} & \underline{172.262} \\
    64    & 3.918 & 12.140 & 43.397 & 55.849 & \underline{96.881} & \underline{161.446} & \underline{156.240} & \underline{182.860} \\
    \hline
    \end{tabular}%
  \label{tab:r3}%
\end{table}%

\begin{table}[!h]
  \scriptsize
  \centering
  \renewcommand{\arraystretch}{0.9}
  \caption{The average number of search nodes for each Type 2 instance group.}
    \begin{tabular}{rrrrrrrrr}
    \hline
    \multirow{2}[0]{*}{$n$} & \multicolumn{8}{c}{$m$} \\
    \cline{2-9}
       $n$ & 8     & 10    & 12    & 14    & 16    & 18    & 20    & 22 \\
    \hline
    16    & 92    & 132   & 262   & 305   & 511   & 585   & 959   & 1,044 \\
    18    & 173   & 288   & 381   & 579   & 1,051 & 1,650 & 1,920 & 2,490 \\
    20    & 278   & 464   & 797   & 1,248 & 1,559 & 2,969 & 3,763 & 5,081 \\
    22    & 416   & 673   & 1,116 & 2,439 & 3,214 & 5,232 & 7,791 & 10,174 \\
    24    & 480   & 1,300 & 2,104 & 3,521 & 6,002 & 8,872 & 16,824 & 26,084 \\
    26    & 950   & 1,685 & 3,526 & 5,860 & 10,500 & 23,521 & 33,464 & 42,133 \\
    28    & 1,014 & 3,504 & 8,916 & 15,068 & 18,140 & 33,917 & 66,739 & 103,750 \\
    30    & 2,052 & 4,472 & 9,943 & 21,104 & 34,806 & 79,595 & 105,243 & 226,899 \\
    32    & 3,415 & 4,779 & 15,232 & 36,176 & 93,686 & 130,977 & 235,314 & 665,416 \\
    34    & 4,022 & 10,115 & 18,945 & 71,387 & 95,621 & 247,737 & 503,058 & 767,987 \\
    36    & 4,974 & 12,923 & 61,748 & 140,081 & 183,606 & 392,267 & 1,150,811 & 1,266,911 \\
    38    & 3,993 & 29,039 & 43,450 & 196,504 & 313,253 & 928,228 & 1,554,085 & 1,984,941 \\
    40    & 7,703 & 25,478 & 64,107 & 141,726 & 618,351 & 616,794 & 1,728,150 & 2,221,610 \\
    42    & 11,143 & 32,937 & 89,721 & 286,598 & 754,305 & 1,356,438 & 2,120,414 & 4,059,816 \\
    44    & 16,423 & 59,929 & 135,760 & 448,782 & 947,224 & 2,620,935 & 2,794,299 & \underline{3,453,335} \\
    46    & 22,400 & 147,524 & 145,857 & 529,693 & 1,031,659 & 1,684,446 & 3,729,016 & \underline{4,316,409} \\
    48    & 25,743 & 74,511 & 364,220 & 659,308 & 1,953,064 & 2,823,159 & 3,983,337 & \underline{4,556,976} \\
    50    & 29,874 & 85,712 & 567,577 & 1,088,311 & 2,384,690 & 3,159,260 & 4,686,411 & \underline{4,891,403} \\
    52    & 31,840 & 210,741 & 716,255 & 1,979,313 & 1,973,444 & 4,318,443 & \underline{4,747,867} & \underline{5,172,142} \\
    54    & 64,838 & 192,700 & 1,130,890 & 1,208,212 & 3,341,742 & \underline{4,504,014} & \underline{5,936,410} & \underline{6,539,519} \\
    56    & 58,341 & 218,177 & 867,317 & 2,644,071 & 2,493,294 & 4,467,560 & \underline{5,734,008} & \underline{5,948,378} \\
    58    & 87,367 & 346,222 & 1,711,089 & 1,950,019 & 2,943,275 & \underline{5,249,511} & \underline{6,886,448} & \underline{6,844,526} \\
    60    & 100,968 & 402,869 & 1,245,312 & 3,676,459 & 4,677,957 & \underline{6,569,635} & \underline{4,345,421} & \underline{8,538,072} \\
    62    & 64,903 & 546,580 & 1,530,491 & 3,163,571 & \underline{5,765,984} & \underline{6,796,924} & \underline{7,023,583} & \underline{6,804,257} \\
    64    & 166,009 & 655,791 & 2,308,876 & 2,718,102 & \underline{4,356,074} & \underline{6,882,350} & \underline{6,422,940} & \underline{7,377,196} \\
    \hline
    \end{tabular}%
  \label{tab:r4}%
\end{table}%

\begin{figure}[!h]
\[
\begin{array}{cc}
\resizebox{7cm}{!}{\includegraphics{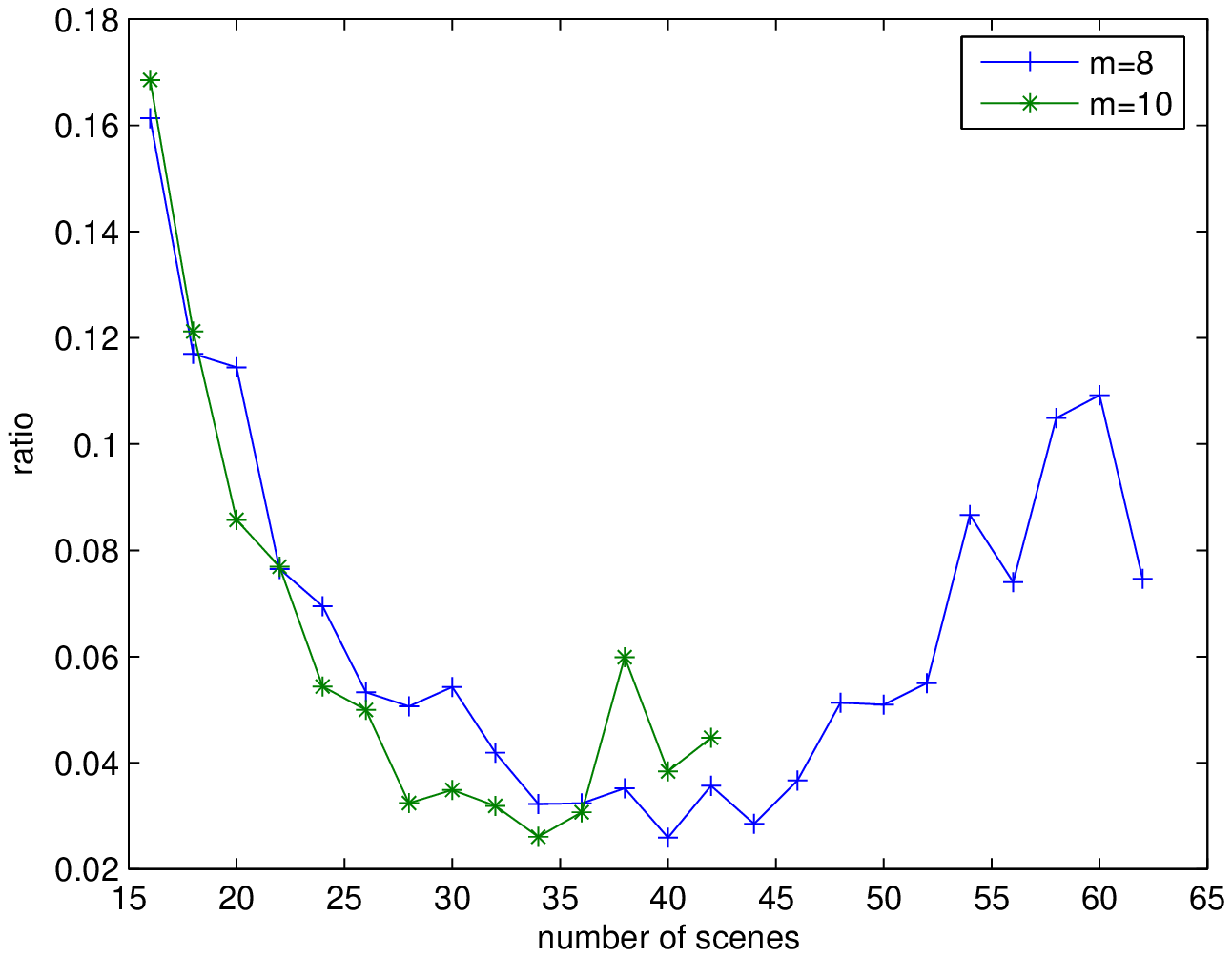}} & \resizebox{7cm}{!}{\includegraphics{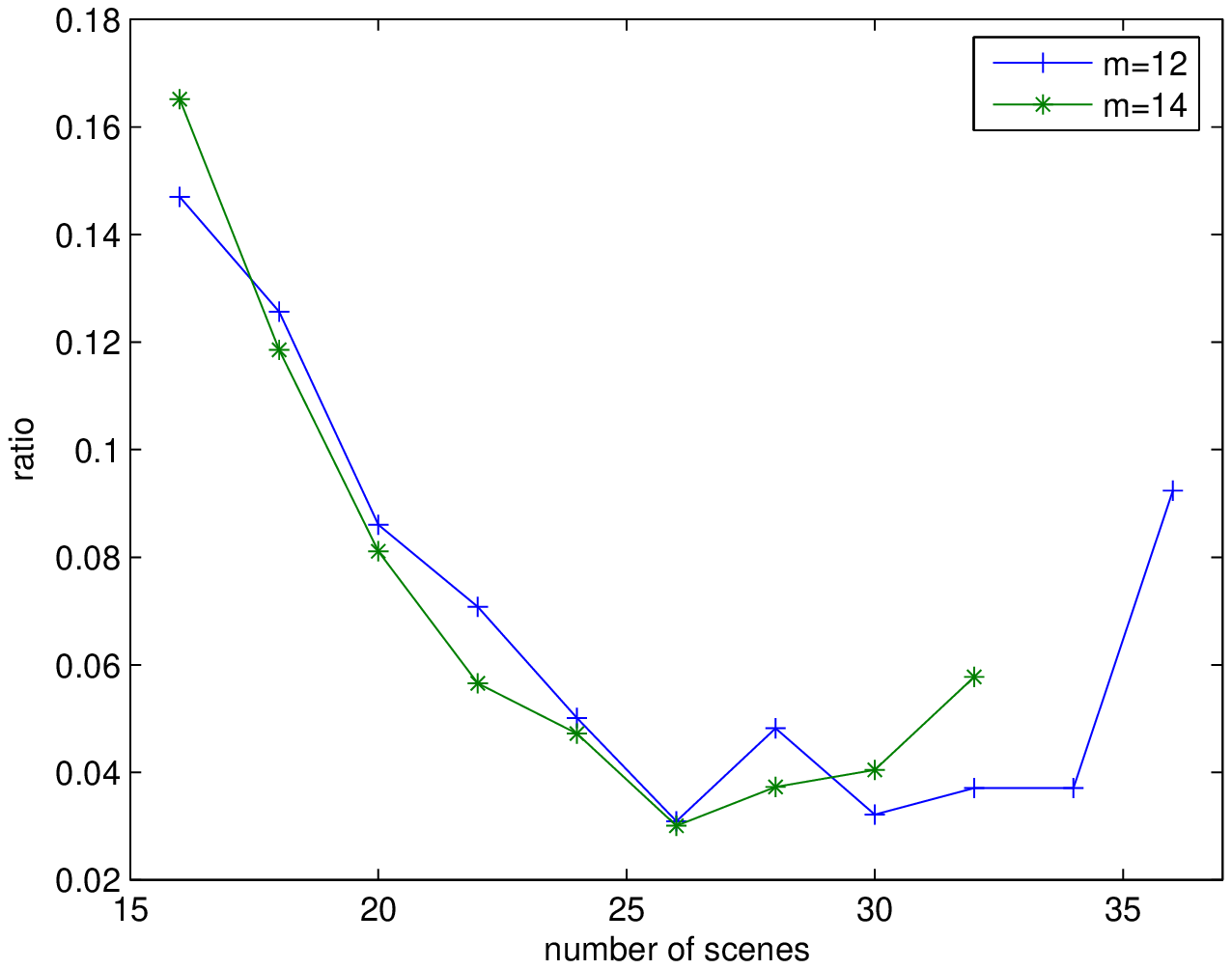}} \\
\mbox{(a) \ \ } & \mbox{(b) \ \ }\\
&\\
\resizebox{7cm}{!}{\includegraphics{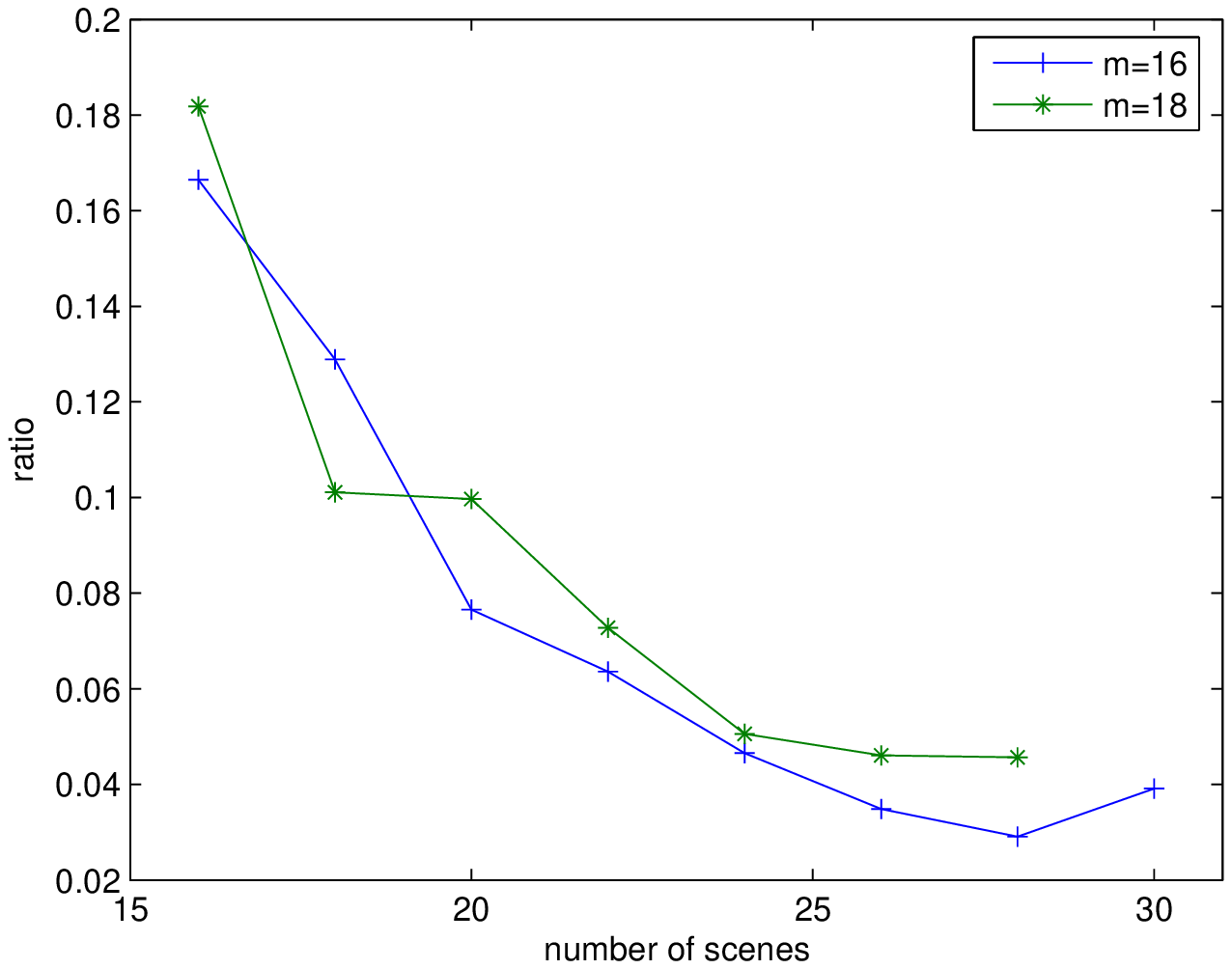}}&\resizebox{7cm}{!}{\includegraphics{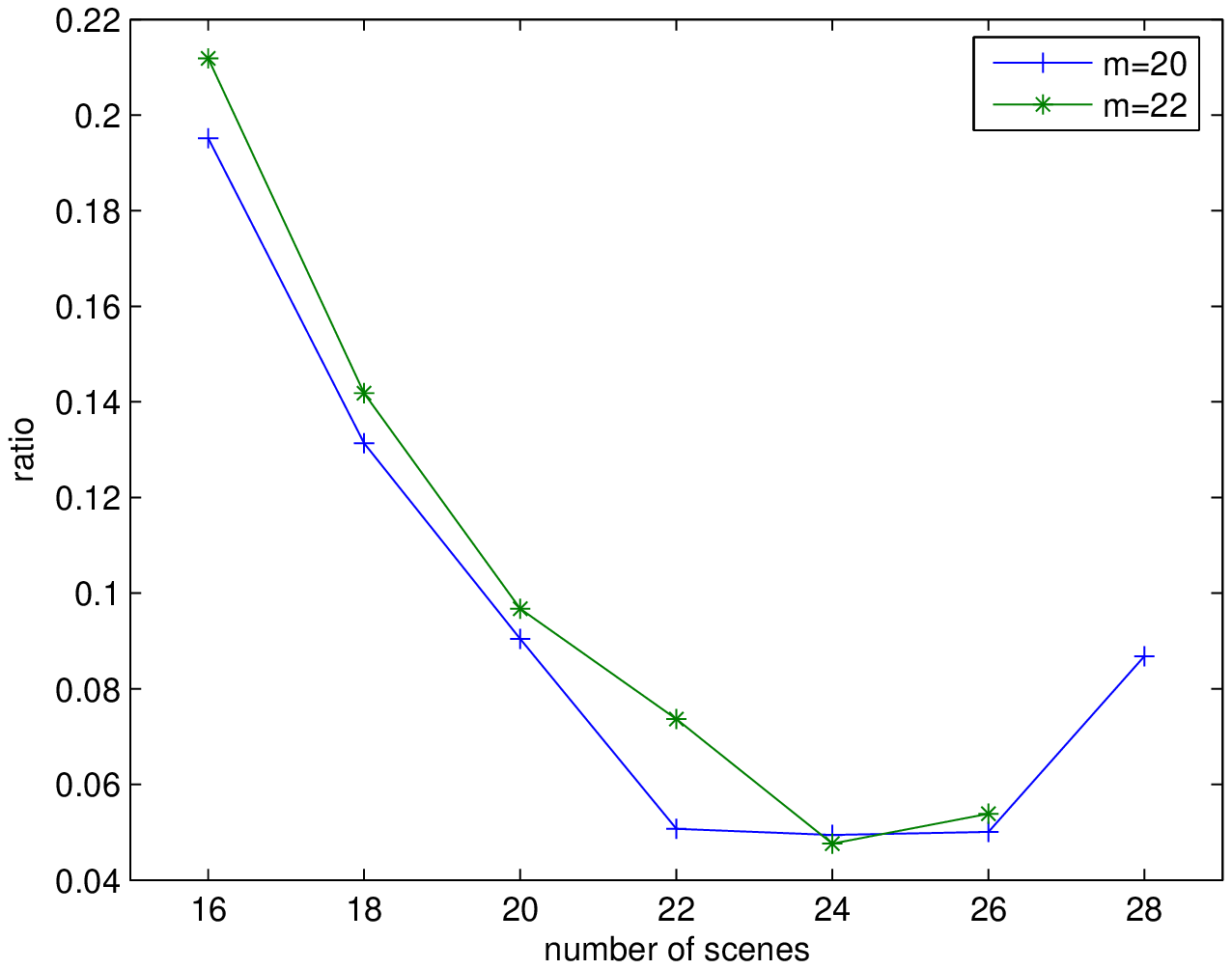}} \\
\mbox{(c) \ \ } & \mbox{(d) \ \ }
\end{array}
\]
\caption{(a) $m = \{8, 10\}$. (b) $m = \{12, 14\}$. (c) $m = \{16, 18\}$. (d) $m = \{20, 22\}$.} \label{fig:4p}
\end{figure}

To further compare our results with those reported by \citet{de2011solving}, we pictorially show in Figure \ref{fig:4p} the ratio of the average number of subproblems (i.e., search nodes) generated by our algorithm to that generated by the dynamic programming algorithm. Each point in these curves corresponds to an instance group whose average number of subproblems was reported by \citet{de2011solving}. On average, the number of subproblems generated by our algorithm is  less than 22\% of that generated by the dynamic programming algorithm, which should be attributed to the use of the new lower bound and domination rules. Moreover, we can observe some trends from these curves. The ratio decreases as the number of scenes increases at the early stage, which implies that our algorithm can eliminate more subproblems. Subsequently, the ratio increases with the number of scenes. This is because hash collisions happened more frequently, reducing the opportunities of pruning search nodes and therefore increasing the number of subproblems.

\subsection{Impacts of Parameter Settings}
\label{sec:imp}
We taken the value of $C$ from $\{0, 2^5, 2^{10}, 2^{15}, 2^{20}, 2^{25}\}$, where $C = 0$ means that cache is not used. Considering the two caching strategies, we have 12 parameter combinations in total. We tested these 12 parameter combinations using a portion of the Type 2 instances. Specifically, the first 5 instances were selected from each instance group, for a total of 1,000 instances. We also imposed a time limit of 10 minutes on each execution of our algorithm. The results of those optimally solved instances were recorded for analysis.

\begin{figure}[!htp]
\begin{center}
\resizebox{10cm}{!}{\includegraphics{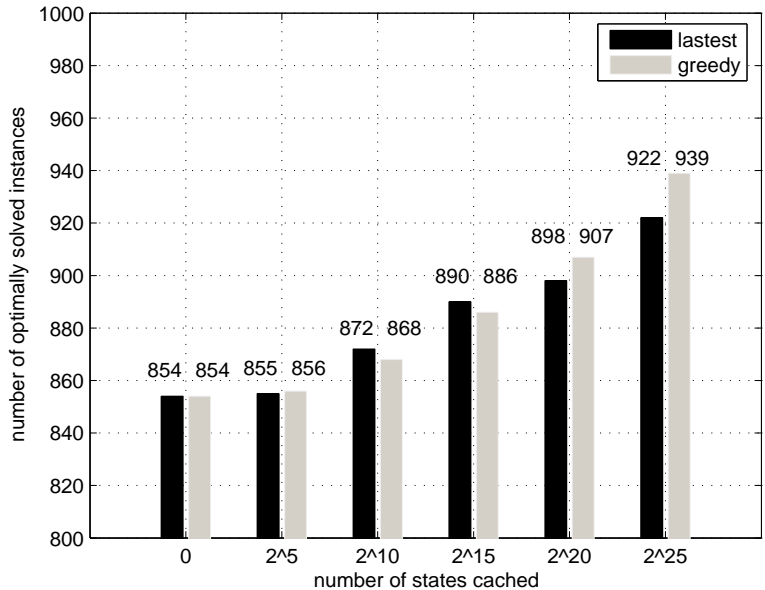}}
\end{center}
\caption{The impact of different parameter settings on the number of optimally solved instances.} \label{fig:3}
\end{figure}

Figure \ref{fig:3} illustrates the number of optimally solved instances under each parameter setting. This figure shows that more caching states lead to more optimally solved instances under both caching strategies. Under the \textit{latest} caching strategy, the number of instances optimally solved increases from 854 ($C=0$) to 922 ($C=2^{25}$). Under the \textit{greedy} caching strategy, this number increases from 854 to 939. When $C$ is relatively small (e.g., $C\leq 2^{15}$), hash collisions occur frequently and the \textit{latest} caching strategy leads to slightly better performance than the \textit{greedy} caching strategy. The {\em greedy} caching strategy may store more states associated with the subproblems at the early level of the search tree, which cannot be used to effectively prune the nodes. We conjecture that since the {\em latest} caching strategy stores the newly encountered states and a certain state is revisited in short period with high probability, the pruning can occur with more opportunities and then the number of subproblems is reduced. When $C$ is large (e.g., $C \geq 2^{20}$), the \textit{greedy} caching strategy leads to more optimally solved instances than the \textit{latest} caching strategy. This may be because a smaller state value in the caching slot is likely to eliminate more subproblems during the search process.

\begin{figure}[!h]
\[
\begin{array}{cc}
\resizebox{8cm}{!}{\includegraphics{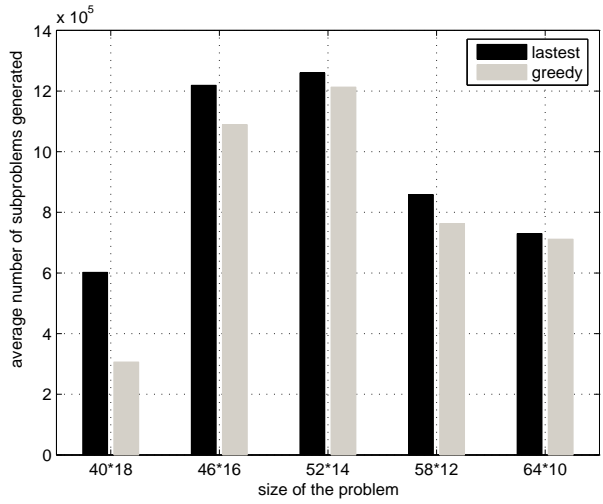}} & \resizebox{8cm}{!}{\includegraphics{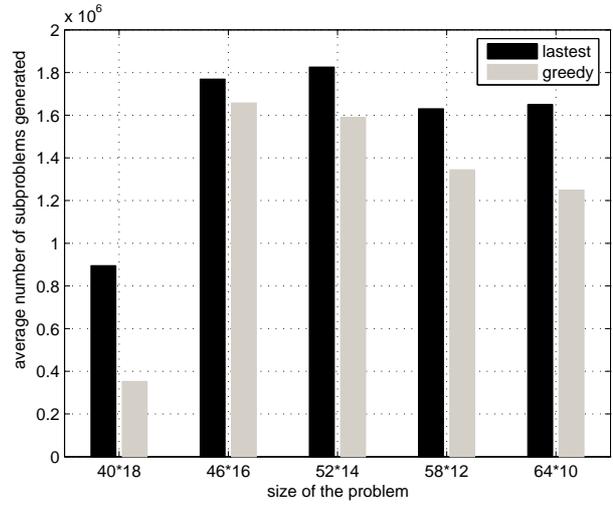}} \\
\mbox{(a) \ \ } & \mbox{(b) \ \ }\\
&\\
\resizebox{7.7cm}{!}{\includegraphics{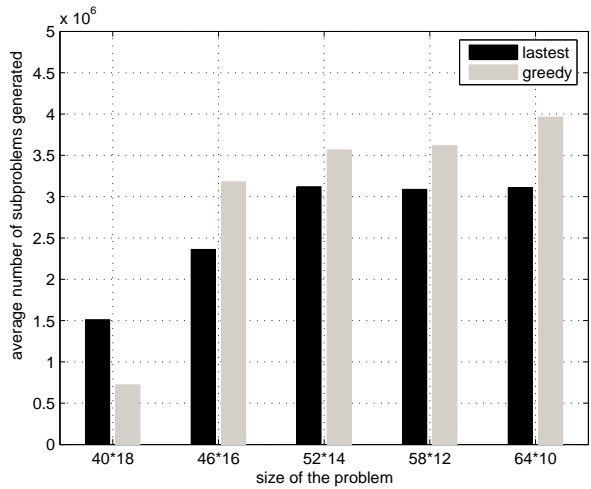}}&\resizebox{7.7cm}{!}{\includegraphics{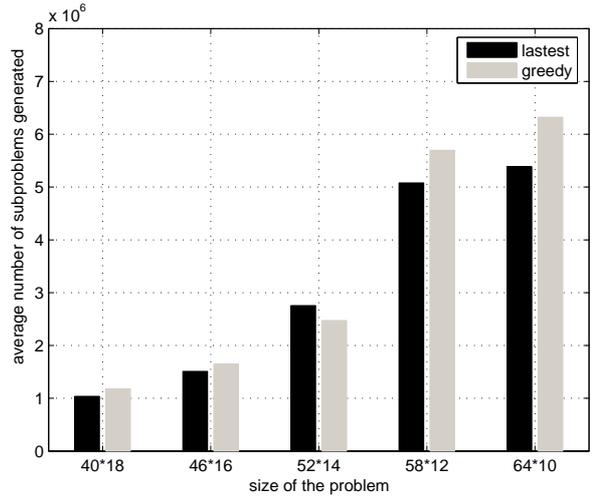}} \\
\mbox{(c) \ \ } & \mbox{(d) \ \ }
\end{array}
\]
%\caption{(a) $C = 2^{25}$. (b) $C = 2^{20}$. (c) $C = 2^{15}$. (d) $C = 2^{10}$.} \label{fig:4}
\caption{(a) $C = 2^{10}$. (b) $C = 2^{15}$. (c) $C = 2^{20}$. (d) $C = 2^{25}$.} \label{fig:4}
\end{figure}

To further test the impacts of different parameter settings on the average number of subproblems generated, we selected five Type 2 instance groups, namely $40 \times 18$, $46 \times 16$, $52 \times 14$, $58 \times 12$ and $64 \times 10$. All instances in these five groups can be optimally solved using our branch-and-bound algorithm within 10 minutes of running time. We pictorially show the results associated with some parameter settings in Figure \ref{fig:4}. We can clearly observe that the average number of subproblems generated decreases as the number of cached states increases. This is in accordance with our intuition since more cache slots store more states, which helps prune more search nodes and therefore reduces the number of subproblems. This figure also reveals that the \textit{greedy} caching strategy outperforms the \textit{latest} caching strategy in terms of the average number of subproblems generated when $C=2^{20}$ or  $C = 2^{25}$, while the \textit{latest} caching strategy generally generates fewer subproblems when $C$ is small, i.e., $C= 2^{10}$ or $C = 2^{15}$. As a result, we adopted the {\em greedy} caching strategy and $C = 2^{25}$ in the final implementation of our branch-and-bound algorithm.

\section{Conclusions}
\label{sec:con}
In this paper, we proposed an enhanced branch-and-bound algorithm to solve the talent scheduling problem, which is a very challenging combinatorial optimization problem. This algorithm uses a new lower bound and two new dominance rules to prune the search nodes. In addition, it caches search states for the purpose of eliminating search nodes. The experimental results clearly show that our algorithm outperforms the current best approach and achieved the optimal solutions for considerably more benchmark instances.

We present a mixed integer linear programming model for the talent scheduling problem in Section \ref{sec:mf}. A possible future research direction is to design mathematical programming algorithms for the talent scheduling problem, such as branch-and-cut algorithm and branch-and-bound coupled with lagrangian relaxation and sub-gradient methods.

\section*{Acknowledgments}
This research was partially supported by the Fundamental Research Funds for the Central Universities, HUST (Grant No. 2012QN213) and National Natural Science Foundation of China (Grant No. 71201065 and 71131004).

\bibliographystyle{elsarticle-harv}
\bibliography{reference}






% \end{thebibliography}

\end{document}